\RequirePackage{fix-cm}
\documentclass[smallcondensed]{svjour3}     
\smartqed  
\usepackage{graphicx}
\usepackage{mathptmx}      
\usepackage{color}
\usepackage{theapa}
\usepackage{float}
\usepackage{graphicx}
\usepackage{amsmath}

\usepackage{amsthm}
\usepackage{amsfonts}
\usepackage{algpseudocode}
\usepackage[caption=false,font=normalsize,labelfont=sf,textfont=sf]{subfig}
\usepackage{threeparttable}
\usepackage{url}
\DeclareMathOperator*{\argmin}{arg\,min}
\newtheorem{dfn}{Definition}
\newtheorem{rmk}{Remark}
\newtheorem{eg}{Example}
\journalname{Machine Learning}
\begin{document}

\title{Clustering with Missing Features: a Penalized Dissimilarity Measure based Approach
}


\author{Shounak Datta \and
        Supritam Bhattacharjee \and
        Swagatam Das* \thanks{*Corresponding Author} 
}


\institute{S. Datta \and S. Das \at
              Electronics and Communication Sciences Unit, Indian Statistical Institute, 203, B. T. Road, Kolkata-700 108, India. \\
              Tel.: +91-33-2575-2323\\
              \email{swagatam.das@isical.ac.in}           
           \and
           S. Bhattacharjee \at
              formerly at Instrumentation \& Electronics Engineering Department, Jadavpur University Salt Lake Campus, Salt Lake City , Block-LB, Plot No. 8, Sector - III, Kolkata - 700 098, India. \\
}

\date{Received: date / Accepted: date}

\maketitle

\begin{abstract}
Many real-world clustering problems are plagued by incomplete data characterized by missing or absent features for some or all of the data instances. Traditional clustering methods cannot be directly applied to such data without preprocessing by imputation or marginalization techniques. In this article, we overcome this drawback by utilizing a penalized dissimilarity measure which we refer to as the Feature Weighted Penalty based Dissimilarity (FWPD). Using the FWPD measure, we modify the traditional k-means clustering algorithm and the standard hierarchical agglomerative clustering algorithms so as to make them directly applicable to datasets with missing features. We present time complexity analyses for these new techniques and also undertake a detailed theoretical analysis showing that the new FWPD based k-means algorithm converges to a local optimum within a finite number of iterations. We also present a detailed method for simulating random as well as feature dependent missingness. We report extensive experiments on various benchmark datasets for different types of missingness showing that the proposed clustering techniques have generally better results compared to some of the most well-known imputation methods which are commonly used to handle such incomplete data. We append a possible extension of the proposed dissimilarity measure to the case of absent features (where the unobserved features are known to be undefined).

\end{abstract}

\section{Introduction}\label{sec:intro}

In data analytics, clustering is a fundamental technique concerned with partitioning a given dataset into useful groups (called \emph{clusters}) according to the relative similarity among the data instances. Clustering algorithms attempt to partition a set of data instances (characterized by some \emph{features}), into different clusters such that the member instances of any given cluster are akin to each other and are different from the members of the other clusters. Greater the similarity within a group and the dissimilarity between groups, better is the clustering obtained by a suitable algorithm.

\par Clustering techniques are of extensive use and are hence being constantly investigated in statistics, machine learning, and pattern recognition. Clustering algorithms find applications in various fields such as economics, marketing, electronic design, space research, etc. For example, clustering has been used to group related documents for web browsing \cite{broder1997syntactic,haveliwala2000scalable}, by banks to cluster the previous transactions of clients to identify suspicious (possibly fraudulent) behaviour \cite{sabau2012survey}, for formulating effective marketing strategies by clustering customers with similar behaviour \cite{chaturvedi1997feature}, in earthquake studies for identifying dangerous zones based on previous epicentre locations \cite{weatherill2009delineation,shelly2009precise,lei2010identify}, and so on. However, when we analyze such real-world data, we may encounter incomplete data where some features of some of the data instances are missing. For example, web documents may have some expired hyper-links. Such missingness may be due to a variety of reasons such as data input errors, inaccurate measurement, equipment malfunction or limitations, and measurement noise or data corruption, etc. This is known as unstructured missingness \cite{chan1972oldest,rubin1976inference}. Alternatively, not all the features may be defined for all the data instances in the dataset. This is termed as structural missingness or absence of features \cite{chechik2008absent}. For example, credit-card details may not be defined for non-credit card clients of a bank.

\par Missing features have always been a challenge for researchers because traditional learning methods (which assume all data instances to be \emph{fully observed}, i.e. all the features are observed) cannot be directly applied to such incomplete data, without suitable preprocessing. When the rate of missingness is low, the data instances with missing values may be ignored. This approach is known as \emph{marginalization}. Marginalization cannot be applied to data having a sizable number of missing values, as it may lead to the loss of a sizable amount of information. Therefore, sophisticated methods are required to fill in the vacancies in the data, so that traditional learning methods can be applied subsequently. This approach of filling in the missing values is called \emph{imputation}. However, inferences drawn from data having a large fraction of missing values may be severely warped, despite the use of such sophisticated imputation methods \cite{acuna2004treatment}.

\subsection{Literature}\label{sec:lit}

The initial models for feature missingness are due to Rubin and Little \citeyear{rubin1976inference,little1987statistical}. They proposed a three-fold classification of missing data mechanisms, viz. Missing Completely At Random (MCAR), Missing At Random (MAR), and Missing Not At Random (MNAR). MCAR refers to the case where missingness is entirely haphazard, i.e. the likelihood of a feature being unobserved for a certain data instance depends neither on the observed nor on the unobserved characteristics of the instance. For example, in an annual income survey, a citizen is unable to participate, due to unrelated reasons such as traffic or schedule problems. MAR eludes to the cases where the missingness is conditional to the observed features of an instance, but is independent of the unobserved features. Suppose, college-goers are less likely to report their income than office-goers. But, whether a college-goer will report his or her income is independent of the actual income. MNAR is characterized by the dependence of the missingness on the unobserved features. For example, people who earn less are less likely to report their incomes in the annual income survey. Datta et al. \citeyear{datta2016fwpd} further classified MNAR into two sub-types, namely MNAR-I when the missingness only depends on the unobserved features and MNAR-II when the missingness is governed by both observed as well as unobserved features. Schafer \& Graham \citeyear{schafer2002missing} and Zhang et al. \citeyear{zhang2012software} have observed that MCAR is a special case of MAR and that MNAR can also be converted to MAR by appending a sufficient number of additional features. Therefore, most learning techniques are based on the validity of the MAR assumption.

\par A lot of research on the problem of learning with missing or absent features has been conducted over the past few decades, mostly focussing on imputation methods. Several works such as \cite{little1987statistical} and \cite{schafer1997analysis} provide elaborate theories and analyses of missing data. Common imputation methods \cite{donders2006review} involve filling the missing features of data instances with zeros (Zero Imputation (ZI)), or the means of the corresponding features over the entire dataset (Mean Imputation (MI)). Class Mean Imputation or Concept Mean Imputation (CMI) is a slight modification of MI that involves filling the missing features with the average of all observations having the same label as the instance being filled. Yet another common imputation method is k-Nearest Neighbor Imputation (kNNI) \cite{dixon1979pattern}, where the missing features of a data instance are filled in by the means of corresponding features over its k-Nearest Neighbors (kNN), on the observed subspace. 
Grzymala-Busse \& Hu \citeyear{grzymala2001comparison} suggested various novel imputation schemes such as treating missing attribute values as special values. 
Rubin \citeyear{rubin1987multiple} proposed a technique called Multiple Imputation (MtI) to model the uncertainty inherent in imputation. In MtI, the missing values are imputed by a typically small (e.g. 5-10) number of simulated versions, depending on the percentage of missing data \cite{chen2013nomore,horton2001multiple}. 
Some more sophisticated imputation techniques have been developed, especially by the bioinformatics community, to impute the missing values by exploiting the correlations between data. A prominent examples is the Singular Value Decomposition based Imputation (SVDI) technique \cite{troyanskaya2001missing} which performs regression based estimation of the missing values using the k most significant eigenvectors of the dataset. Other examples inlcude Least Squares Imputation (LSI) \cite{bo2004lsimpute}, Non-Negative LSI (NNLSI) and Collateral Missing Value Estimation (CMVE) \cite{sehgal2005collateral}. Model-based methods are related to yet distinct from imputation techniques. These methods attempt to model the distributions for the missing values instead of filling them in \cite{dempster1983incomplete,ahmad1993some,wang2002empirical1,wang2002empirical2}.


\par However, most of these techniques assume the pattern of missingness to be MCAR or MAR because this allows the use of simpler models of missingness \cite{heitjan1996distinguishing}. Such simple models are not likely to perform well in case of MNAR as the pattern of missingness also holds information. Hence, other methods have to be developed to tackle incomplete data due to MNAR \cite{marlin2008missing}. Moreover, imputation may often lead to the introduction of noise and uncertainty in the data \cite{dempster1983incomplete,little1987statistical,barcelo2008impact,myrtveit2001analyzing}. 

\par In light of the observations made in the preceding paragraph, some learning methods avoid the inexact methods of imputation (as well as marginalization) altogether, while dealing with missingness. A common paradigm is random subspace learning where an ensemble of learners is trained on projections of the data in random subspaces and an inference is drawn based on the concensus among the ensemble \cite{krause2003ensemble,juszczak2004combining,nanni2012classifier}. Chechik et al. \citeyear{chechik2008absent} used the geometrical insight of max-margin classification to formulate an objective function which was optimized to directly classify the incomplete data. This was extended to the max-margin regression case for software effort prediction with absent features in \cite{zhang2012software}. Wagstaff et al. \citeyear{wagstaff2004clustering,wagstaff2005making} suggested a k-means algorithm with Soft Constraints (KSC) where soft constraints determined by fully observed objects are introduced to facilitate the grouping of instances with missing features. 
Himmelspach \& Conrad \citeyear{himmelspach2010clustering} provided a good review of partitional clustering techniques for incomplete datasets, which mentions some other techniques that do not make use of imputation.

\par The idea to modify the distance between the data instances to directly tackle missingness (without having to resort to imputation) was first put forth by Dixon \citeyear{dixon1979pattern}. The Partial Distance Strategy (PDS) proposed in \cite{dixon1979pattern} scales up the \emph{observed distance}, i.e. the distance between two data instances in their \emph{common observed subspace} (the subspace consisting of the observed features common to both data instances) by the ratio of the total number of features (observed as well as unobserved) and the number of common observed features between them to obtain an estimate of their distance in the fully observed space. Hathaway \& Bezdek \citeyear{hathaway2001fcm} used the PDS to extend the Fuzzy C-Means (FCM) clustering algorithm to cases with missing features. Furthermore, Mill{\'a}n-Giraldo et al. \citeyear{millan2010dissimilarity} and Porro-Mu{\~n}oz et al. \citeyear{porro2013missing} generalized the idea of the PDS by proposing to scale the observed distance by factors other than the fraction of observed features. However, neither the PDS nor its extensions can always provide a good estimate of the actual distance as the observed distance between two instances may be unrelated to the distance between them in the unobserved subspace.

\subsection{Motivation}\label{sec:motiv}

As observed earlier, one possible way to adapt supervised as well as unsupervised learning methods to problems with missingness is to modify the \emph{distance} or \emph{dissimilarity measure} underlying the learning method. The idea is that the modified dissimilarity measure should use the common observed features to provide approximations of the distances between the data instances if they were to be fully observed. PDS is one such measure. Such approaches neither require marginalization nor imputation and are likely to yield better results than either of the two. For example, let $X_{full}=\{\mathbf{x}_1=(1,2),\mathbf{x}_2=(1.8,1),\mathbf{x}_3=(2,2.5)\}$ be a dataset consisting of three points in $\mathbb{R}^2$. Then, we have $d_{E}(\mathbf{x}_1,\mathbf{x}_2)=1.28$ and $d_{E}(\mathbf{x}_1,\mathbf{x}_3)=1.12$, $d_{E}(\mathbf{x}_i,\mathbf{x}_j)$ being the Euclidean distance between any two fully observed points $\mathbf{x}_i$ and $\mathbf{x}_j$ in $X_{full}$. Suppose that the first coordinate of the point $(1,2)$ be unobserved, resulting in the incomplete dataset $X=\{\widetilde{\mathbf{x}}_1=(*,2),\mathbf{x}_2=(1.8,1),\mathbf{x}_3=(2,2.5)\}$ (`*' denotes a missing value), on which learning must be carried out. Notice that this is a case of unstructured missingness (because the unobserved value is known to exist), as opposed to the structural missingness of \cite{chechik2008absent}. Using ZI, MI and 1NNI respectively, we obtain the following filled in datasets:
\begin{equation*}
\begin{aligned}
&X_{ZI} = \{\widehat{\mathbf{x}_1}=(0,2),\mathbf{x}_2=(1.8,1),\mathbf{x}_3=(2,2.5)\},\\
&X_{MI} = \{\widehat{\mathbf{x}_1}=(1.9,2),\mathbf{x}_2=(1.8,1),\mathbf{x}_3=(2,2.5)\},\\
\text{and } &X_{1NNI} = \{\widehat{\mathbf{x}_1}=(2,2),\mathbf{x}_2=(1.8,1),\mathbf{x}_3=(2,2.5)\},
\end{aligned}
\end{equation*}
where $\widehat{\mathbf{x}_1}$ denotes an estimate of $\mathbf{x}_1$. If PDS is used to estimate the corresponding distances in $X$, then the distance $d_{PDS}(\mathbf{x}_1,\mathbf{x}_i)$ between the implicit estimate of $\mathbf{x}_1$ and some other instance $\mathbf{x}_i \in X$ is obtained by
\begin{equation*}
d_{PDS}(\mathbf{x}_1,\mathbf{x}_i) = \sqrt{\frac{2}{1}(x_{1,2}-x_{i,2})^2},
\end{equation*}
where $x_{1,2}$ and $x_{i,2}$ respectively denote the 2nd features of $\mathbf{x}_1$ and $\mathbf{x}_i$, and 2 is the numerator of the multiplying factor due to the fact that $\mathbf{x}_i \in \mathbb{R}^2$ and 1 is the denominator owing to the fact that only the 2nd feature is observed for both $\mathbf{x}_1$ and $\mathbf{x}_i$. Then, we get
\begin{equation*}
\begin{aligned}
&d_{PDS}(\mathbf{x}_1,\mathbf{x}_2) = \sqrt{\frac{2}{1}(2-1)^2} = 1.41,\\
\text{and } &d_{PDS}(\mathbf{x}_1,\mathbf{x}_3) = \sqrt{\frac{2}{1}(2-2.5)^2} = 0.71.
\end{aligned}
\end{equation*}
The improper estimates obtained by PDS are due to the fact that the distance in the common observed subspace does not reflect the distance in the unobserved subspace. This is the principal drawback of the PDS method, as discussed earlier. Since the observed distance between two data instances is essentially a lower bound on the Euclidean distance between them (if they were to be fully observed), adding a suitable penalty to this lower bound can result in a reasonable approximation of the actual distance. This approach \cite{datta2016fwpd} called the Penalized Dissimilarity Measure (PDM) may be able to overcome the drawback which plagues PDS. Let the penalty between $\mathbf{x}_1$ and $\mathbf{x}_i$ be given by the ratio of the number of features which are unobserved for at least one of the two data instances and the total number of features in the entire dataset. Then, the dissimilarity $\delta_{PDM}(\mathbf{x}_1,\mathbf{x}_i)$ between the implicit estimate of $\mathbf{x}_1$ and some other $\mathbf{x}_i \in X$ is
\begin{equation*}
\delta_{PDM}(\mathbf{x}_1,\mathbf{x}_i) = \sqrt{(x_{1,2}-x_{i,2})^2} + \frac{1}{2},
\end{equation*}
where the 1 in the numerator of the penalty term is due to the fact that the 1st feature of $\mathbf{x}_1$ is unobserved. Therefore, the dissimilarities $\delta_{PDM}(\mathbf{x}_1,\mathbf{x}_2)$ and $\delta_{PDM}(\mathbf{x}_1,\mathbf{x}_3)$ are
\begin{equation*}
\begin{aligned}
&\delta_{PDM}(\mathbf{x}_1,\mathbf{x}_2) = \sqrt{(2-1)^2} + \frac{1}{2} = 1.5,\\
\text{and } &\delta_{PDM}(\mathbf{x}_1,\mathbf{x}_3) = \sqrt{(2-2.5)^2} + \frac{1}{2} = 1.
\end{aligned}
\end{equation*}




\begin{figure}[!t]
\centering
\subfloat[Comparison for $\mathbf{x}_1$.]{\includegraphics[width=2.3in]{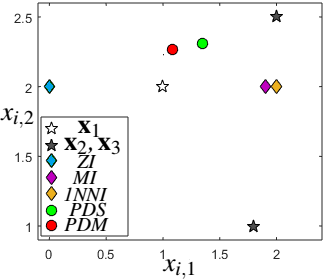}
\label{figVariousMissing1}}
\hfill
\subfloat[Comparison for $\mathbf{x'}_1$.]{\includegraphics[width=2.27in]{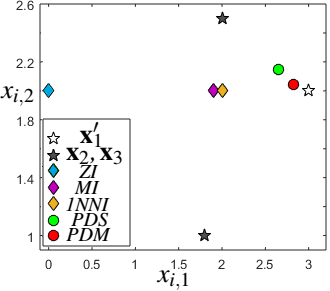}
\label{figVariousMissing2}}
\caption{Comparison of various techniques for handling missing features.}
\label{figVarious}
\end{figure}

\par The situation is illustrated in Figure \ref{figVariousMissing1}. The reader should note that the points estimated using ZI, MI and 1NNI exist in the same 2-D Cartesian space to which $X_{full}$ is native. On the other hand, the points estimated by both PDS and PDM exist in their individual abstract spaces (likely distinct from the native 2-D space). Therefore, for the sake of easy comparison, we illustrate all the estimates together by superimposing both these abstract spaces on the native 2-D space so as to coincide at the points $\mathbf{x}_2$ and $\mathbf{x}_3$. It can be seen that the approach based on the PDM does not suffer from the drawback of PDS and is better able to preserve the relationship between the points. Moreover, it should be noted that there are two possible images for each of the estimates obtained by both PDS and PDM. Therefore, had the partially observed point instead been $\mathbf{x'}_1 = (3,2)$ with the first feature missing (giving rise to the same incomplete dataset $X$; $\widetilde{\mathbf{x'}}_1$ replacing the identical incomplete point $\widetilde{\mathbf{x}}_1$), PDS and PDM would still find reasonably good estimates (PDM still being better than PDS). This situation is also illustrated in Figure \ref{figVariousMissing2}. In general,
\begin{enumerate}
\item ZI works well only for missing values in the vicinity of the origin and is also origin dependent;
\item MI works well only when the missing value is near the observed mean of the missing feature;
\item kNNI is reliant on the assumption that neighbors have similar features, but suffers from the drawbacks that missingness may give rise to erroneous neighbor selection and that the estimates are restricted to the range of observed values of the feature in question;
\item PDS suffers from the assumption that the common observed distances reflect the unobserved distances; and
\item none of these methods differentiate between identical incomplete points, i.e. $\widetilde{\mathbf{x}}_1$ and $\widetilde{\mathbf{x}}'_1$ are not differentiated between.
\end{enumerate}
However, a PDM successfully steers clear of all these drawbacks (notice that $\delta(\mathbf{x}_1,\mathbf{x}'_1) = \frac{1}{2}$). Furthermore, such a PDM can also be easily applied to the case of absent features, by slightly modifying the penalty term (see Appendix \ref{apd:first}). This knowledge motivates us to use a PDM to adapt traditional clustering methods to problems with missing features.

\subsection{Contribution}\label{sec:contrib}

The FWPD measure is a PDM used in \cite{datta2016fwpd} for kNN classification of datasets with missing features\footnote{The work of \citeauthor{datta2016fwpd} \citeyear{datta2016fwpd} is based on the FWPD measure originally proposed in the archived version of the current article \cite{DattaBD16}.}. The FWPD between two data instances is a weighted sum of two terms; the first term being the observed distance between the instances and the second being a penalty term. The penalty term is a sum of the penalties corresponding to each of the features which are missing from at least one of the data instances; each penalty being directly proportional to the probability of its corresponding feature being observed. Such a weighting scheme imposes greater penalty if a feature which is observed for a large fraction of the data is missing for a particular instance. On the other hand, if the missing feature is unobserved for a large fraction of the data, then a smaller penalty is imposed.

\par The contributions of the current article are in order:
\begin{enumerate}

\item In the current article, we formulate the k-means clustering problem for datasets with missing features based on the proposed FWPD and develop an algorithm to solve the new formulation. 
\item We prove that the proposed algorithm is guaranteed to converge to a locally optimal solution of the modified k-means optimization problem formulated with the FWPD measure. 
\item We also propose Single Linkage, Average Linkage, and Complete Linkage based HAC methods for datasets plagued by missingness, based on the proposed FWPD. 
\item We provide an extensive discussion on the properties of the FWPD measure. The said discussion is more thorough compared to that of \cite{datta2016fwpd}.
\item We further provide a detailed algorithm for simulating the four types of missingness enumerated in \cite{datta2016fwpd}, namely MCAR, MAR, MNAR-I (missingness only depends on the unobserved features) and MNAR-II (missingness depends on both observed as well as unobserved features).
\item Moreover, since this work presents an alternative to imputation and can be useful in scenarios where imputation is not practical (such as structural missingness), we append an extension of the proposed FWPD to the case of absent features (where the absent features are known to be undefined or non-existent). We also show that the FWPD becomes a semi-metric in the case of structural missingness.

\end{enumerate}

\par Experiments are reported on diverse datasets and covers all four types of missingness. The results are compared with the popularly used imputation techniques. The comparative results indicate that our approach generally achieves better performance than the common imputation approaches used to handle incomplete data.

\subsection{Organization}\label{sec:organiz}

The rest of this paper is organized in the following way. In Section \ref{sec:fwpd}, we elaborate on the properties of the FWPD measure. The next section (Section \ref{sec:Kmeans}) presents a formulation of the k-means clustering problem which is directly applicable to datasets with missing features, based on the FWPD discussed in Section \ref{sec:fwpd}. This section also puts forth an algorithm to solve the optimization problem posed by this new formulation. The subsequent section (Section \ref{sec:Hier}) covers the HAC algorithm formulated using FWPD to be directly applicable to incomplete datasets. Experimental results (based on the missingness simulating mechanism discussed in the same section) are presented in Section \ref{sec:ExpRes}. Relevant conclusions are drawn in Section \ref{sec:conclsn}. Subsequently, Appendix \ref{apd:first} deals with the extension of the proposed FWPD to the case of absent features (structural missingness).

\section{Feature Weighted Penalty based Dissimilarity Measure for Datasets with Missing Features}\label{sec:fwpd}

Let the dataset $X \subset \mathbb{R}^m$, i.e. the data instances in $X$ are each characterized by $m$ feature values in $\mathbb{R}$. Further, let $X$ consist of $n$ instances $\mathbf{x}_i$ ($i \in \{1, 2, \cdots, n\}$), some of which have missing features. Let $\gamma_{\mathbf{x}_i}$ denote the set of observed features for the data point $\mathbf{x}_i$.  Then, the set of all features $S=\bigcup_{i=1}^{n}\gamma_{\mathbf{x}_i}$ and $|S|=m$. The set of features which are observed for all data instances in $X$ is defined as $\gamma_{obs}=\bigcap_{i=1}^{n}\gamma_{\mathbf{x}_i}$. $|\gamma_{obs}|$ may or may not be non-zero. $\gamma_{miss}=S\backslash\gamma_{obs}$ is the set of features which are unobserved for at least one data point in $X$. The important notations used in this section (and beyond) are summarized in Table \ref{tabNota1}.

\begin{table}[h]
\begin{center}
\caption{Some important notations used in Section \ref{sec:fwpd} and beyond.}\label{tabNota1}
\begin{tabular}{c | l}
\hline
Notation & Meaning\\
\hline
$X$ & Dataset with incomplete data points\\
$n$ & Number of data points in $X$\\
$\mathbf{x}_i$ & A data point in $X$\\
$x_{i,l}$ & $l$-th feature of $\mathbf{x}_i$\\
$S$ & Set of all features in $X$\\
$m$ & Number of features in $S$, i.e. $|S|$\\
$\gamma$ & General notation for a set of features in $S$\\
$\gamma_{\mathbf{x}_i}$ & Set of features observed for point $\mathbf{x}_i$\\
$\gamma_{obs}$ & Set of features observed for all instances in $X$\\
$\gamma_{miss}$ & Set of features which are unobserved for some point in $X$\\
$d_{\gamma}(\mathbf{x}_i,\mathbf{x}_j)$ & Distance between poinst $\mathbf{x}_i$ and $\mathbf{x}_j$ in the subspace defined by the features in $\gamma$\\
$d(\mathbf{x}_i,\mathbf{x}_j)$ & Observed distance between points $\mathbf{x}_i$ and $\mathbf{x}_j$\\
$d_E(\mathbf{x}_i,\mathbf{x}_j)$ & Euclidean distance between fully observed points $\mathbf{x}_i$ and $\mathbf{x}_j$\\
$w_l$ & Number of instances in $X$ having observed values for the $l$-th feature\\
$p(\mathbf{x}_i,\mathbf{x}_j)$ & Feature Weighted Penalty (FWP) between $\mathbf{x}_i$ and $\mathbf{x}_j$\\
$p_{\gamma}$ & FWP corresponding to the subspace defined by $\gamma$\\
$\delta(\mathbf{x}_i,\mathbf{x}_j)$ & Feature Weighted Penalty based Dissimilarity (FWPD) between $\mathbf{x}_i$ and $\mathbf{x}_j$\\
$d_{max}$ & Maximum observed distance between any two data points in $X$\\
$\alpha$ & Coefficient of relative importance between observed distance and FWP for FWPD\\
$\rho_{i,j,k}$ & $p(\mathbf{x}_i,\mathbf{x}_j) + p(\mathbf{x}_j,\mathbf{x}_k) - p(\mathbf{x}_k,\mathbf{x}_i)$ for some $\mathbf{x}_i, \mathbf{x}_j, \mathbf{x}_k \in X$\\
$\phi$ & An empty set\\
\hline
\end{tabular}
\end{center}
\end{table}

\begin{dfn}\label{defLowerDist}
Let the distance between any two data instances $\mathbf{x}_i,\mathbf{x}_j \in X$ in a subspace defined by $\gamma$ be denoted as $d_{\gamma}(\mathbf{x}_i,\mathbf{x}_j)$. Then the observed distance (distance in the common observed subspace) between these two points can be defined as
\begin{equation}\label{eqnLowerDist}
d_{\gamma_{\mathbf{x}_i}\bigcap\gamma_{\mathbf{x}_j}}(\mathbf{x}_i,\mathbf{x}_j)=\sqrt{\sum_{l \in \gamma_{\mathbf{x}_i}\bigcap\gamma_{\mathbf{x}_j}}(x_{i,l}-x_{j,l})^2},
\end{equation}
where $x_{i,l}$ denotes the $l$-th feature of the data instance $\mathbf{x}_i$. For the sake of convenience, $d_{\gamma_{\mathbf{x}_i}\bigcap\gamma_{\mathbf{x}_j}}(\mathbf{x}_i,\mathbf{x}_j)$ is simplified to $d(\mathbf{x}_i,\mathbf{x}_j)$ in the rest of this paper.
\end{dfn}

\begin{dfn}\label{defEDist}
If both $\mathbf{x}_i$ and $\mathbf{x}_j$ were to be fully observed, the Euclidean distance $d_{E}(\mathbf{x}_i,\mathbf{x}_j)$ between $\mathbf{x}_i$ and $\mathbf{x}_j$ would be defined as
\begin{equation*}
d_{E}(\mathbf{x}_i,\mathbf{x}_j)=\sqrt{\sum_{l \in S}(x_{i,l}-x_{j,l})^2}.
\end{equation*}
\end{dfn}
Now, since $(\gamma_{\mathbf{x}_i}\cap\gamma_{\mathbf{x}_j})\subseteq S$, and $(x_{i,l}-x_{j,l})^2 \geq 0$ $\forall$ $l \in S$, it follows that
\begin{equation*}
d(\mathbf{x}_i,\mathbf{x}_j) \leq d_{E}(\mathbf{x}_i,\mathbf{x}_j) \text{ } \forall \text{ } \mathbf{x}_i,\mathbf{x}_j \in X.
\end{equation*}
Therefore, to compensate for the distance in the unobserved subspace, we add a Feature Weighted Penalty (FWP) $p(\mathbf{x}_i,\mathbf{x}_j)$ (defined below) to $d(\mathbf{x}_i,\mathbf{x}_j)$.

\begin{dfn}\label{defFwp}
The FWP between $\mathbf{x}_i$ and $\mathbf{x}_j$ is defined as
\begin{equation}\label{eqnDefFwp}
p(\mathbf{x}_i,\mathbf{x}_j)=\frac{\underset{l \in S \backslash (\gamma_{\mathbf{x}_i}\bigcap \gamma_{\mathbf{x}_j})}{\sum}\;w_l}{\underset{l' \in S}{\sum}\;w_{l'}},
\end{equation}
where $w_l \in (0,n]$ is the number of instances in $X$ having observed values of the feature $l$. It should be noted that FWP exacts greater penalty for unobserved occurrences of those features which are observed for a large fraction of the data instances. Moreover, since the value of the FWP solely depends on the taxable subspace $S \backslash (\gamma_{\mathbf{x}_i}\bigcap \gamma_{\mathbf{x}_j})$, we define an alternative notation for the FWP, viz. $p_{\gamma}={\underset{l \in \gamma}{\sum}\;w_l}/{\underset{l' \in S}{\sum}\;w_{l'}}$. Hence, $p(\mathbf{x}_i,\mathbf{x}_j)$ can also be written as $p_{S \backslash (\gamma_{\mathbf{x}_i}\bigcap \gamma_{\mathbf{x}_j})}$.
\end{dfn}

Then, the definition of the proposed FWPD follows.
\begin{dfn}\label{defFwpd}
The FWPD between $\mathbf{x}_i$ and $\mathbf{x}_j$ is
\begin{equation}\label{eqnFwpd}
\delta(\mathbf{x}_i,\mathbf{x}_j)=(1-\alpha)\times \frac{d(\mathbf{x}_i,\mathbf{x}_j)}{d_{max}} + \alpha \times p(\mathbf{x}_i,\mathbf{x}_j),
\end{equation}
where $\alpha \in (0,1]$ is a parameter which determines the relative importance between the two terms and $d_{max}$ is the maximum observed distance between any two points in $X$ in their respective common observed subspaces.
\end{dfn}

\subsection{Properties of the proposed FWPD}\label{sec:propFwpd}

In this subsection, we discuss some of the important properties of the proposed FWPD measure. The following theorem discusses some of the important properties of the proposed FWPD measure and the subsequent discussion is concerned with the triangle inequality in the context of FWPD.
\begin{theorem}\label{thmPropFwpd}
The FWPD measure satisfies the following important properties:
\begin{enumerate}
\item $\delta(\mathbf{x}_i,\mathbf{x}_i) \leq \delta(\mathbf{x}_i,\mathbf{x}_j)$ $\forall$ $\mathbf{x}_i,\mathbf{x}_j \in X$,
\item $\delta(\mathbf{x}_i,\mathbf{x}_i) \geq 0$ $\forall$ $\mathbf{x}_i \in X$,
\item $\delta(\mathbf{x}_i,\mathbf{x}_i) = 0$ iff $\gamma_{\mathbf{x}_i}=S$, and
\item $\delta(\mathbf{x}_i,\mathbf{x}_j)=\delta(\mathbf{x}_j,\mathbf{x}_i)$ $\forall$ $\mathbf{x}_i,\mathbf{x}_j \in X$.
\end{enumerate}

\begin{proof}\label{pfPropFwpd}
\begin{enumerate}
\item From Equations (\ref{eqnLowerDist}) and (\ref{eqnFwpd}), it follows that
    \begin{equation}\label{eqnFwpdSelf}
    \delta(\mathbf{x}_i,\mathbf{x}_i) = \alpha \times p(\mathbf{x}_i,\mathbf{x}_i).
    \end{equation}
    It also follows from Equation (\ref{eqnDefFwp}) that $p(\mathbf{x}_i,\mathbf{x}_i) \leq p(\mathbf{x}_i,\mathbf{x}_j)$ $\forall$ $\mathbf{x}_i,\mathbf{x}_j \in X$. Therefore, $\delta(\mathbf{x}_i,\mathbf{x}_i) \leq \alpha \times p(\mathbf{x}_i,\mathbf{x}_j)$. Since $\alpha \leq 1$, we have $\alpha \times p(\mathbf{x}_i,\mathbf{x}_j) \leq p(\mathbf{x}_i,\mathbf{x}_j)$. Now, it follows from Equation (\ref{eqnFwpd}) that $p(\mathbf{x}_i,\mathbf{x}_j) \leq \delta(\mathbf{x}_i,\mathbf{x}_j)$. Hence, we get $\delta(\mathbf{x}_i,\mathbf{x}_i) \leq \delta(\mathbf{x}_i,\mathbf{x}_j)$ $\forall$ $\mathbf{x}_i,\mathbf{x}_j \in X$.
\item It can be seen from Equation (\ref{eqnFwpd}) that $\delta(\mathbf{x}_i,\mathbf{x}_i) = \alpha \times p(\mathbf{x}_i,\mathbf{x}_i)$. Moreover, it follows from Equation (\ref{eqnDefFwp}) that $p(\mathbf{x}_i,\mathbf{x}_i) \geq 0$. Hence, $\delta(\mathbf{x}_i,\mathbf{x}_i) \geq 0$ $\forall$ $\mathbf{x}_i \in X$.
\item It is easy to see from Equation (\ref{eqnDefFwp}) that $p(\mathbf{x}_i,\mathbf{x}_i)=0$ iff $\gamma_{\mathbf{x}_i}=S$. Hence, it directly follows from Equation (\ref{eqnFwpdSelf}) that $\delta(\mathbf{x}_i,\mathbf{x}_i) = 0$ iff $\gamma_{\mathbf{x}_i}=S$.
\item From Equation (\ref{eqnFwpd}) we have
    \begin{equation*}
    \begin{aligned}
    & \delta(\mathbf{x}_i,\mathbf{x}_j)=(1-\alpha)\times \frac{d(\mathbf{x}_i,\mathbf{x}_j)}{d_{max}} + \alpha \times p(\mathbf{x}_i,\mathbf{x}_j),\\
    \text{and } & \delta(\mathbf{x}_j,\mathbf{x}_i)=(1-\alpha)\times \frac{d(\mathbf{x}_j,\mathbf{x}_i)}{d_{max}} + \alpha \times p(\mathbf{x}_j,\mathbf{x}_i).
    \end{aligned}
    \end{equation*}
    However, $d(\mathbf{x}_i,\mathbf{x}_j)=d(\mathbf{x}_j,\mathbf{x}_i)$ and $p(\mathbf{x}_i,\mathbf{x}_j)=p(\mathbf{x}_j,\mathbf{x}_i)$ $\forall$ $\mathbf{x}_i,\mathbf{x}_j \in X$ (by definition). Therefore, it can be easily seen that $\delta(\mathbf{x}_i,\mathbf{x}_j)=\delta(\mathbf{x}_j,\mathbf{x}_i)$ $\forall$ $\mathbf{x}_i,\mathbf{x}_j \in X$.
\end{enumerate}
\end{proof}
\end{theorem}

The triangle inequality is an important criterion which lends some useful properties to the space induced by a dissimilarity measure. Therefore, the conditions under which FWPD satisfies the said criterion are investigated below. However, it should be stressed that the satisfaction of the said criterion is not essential for the functioning of the clustering techniques proposed in the subsequent text.
\begin{dfn}\label{defTri}
For any three data instances $\mathbf{x}_i, \mathbf{x}_j, \mathbf{x}_k \in X$, the triangle inequality with respect to (w.r.t.) the FWPD measure is defined as
\begin{equation}\label{eqnTri1}
\delta(\mathbf{x}_i,\mathbf{x}_j) + \delta(\mathbf{x}_j,\mathbf{x}_k) \geq \delta(\mathbf{x}_k,\mathbf{x}_i).
\end{equation}
\end{dfn}
The three following lemmas deal with the conditions under which Inequality (\ref{eqnTri1}) will hold.

\begin{lemma}\label{penTri}
For any three data instances $\mathbf{x}_i, \mathbf{x}_j, \mathbf{x}_k \in X$ let $\rho_{i,j,k} = p(\mathbf{x}_i,\mathbf{x}_j) + p(\mathbf{x}_j,\mathbf{x}_k) - p(\mathbf{x}_k,\mathbf{x}_i)$. Then $\rho_{i,j,k} \geq 0$ $\forall$ $\mathbf{x}_i, \mathbf{x}_j, \mathbf{x}_k \in X$.

\begin{proof}
Let us rewrite the penalty term $p(\mathbf{x}_i,\mathbf{x}_j)$ in terms of the spanned subspaces as $p(\mathbf{x}_i,\mathbf{x}_j) = p_{S \backslash (\gamma_{\mathbf{x}_i} \bigcup \gamma_{\mathbf{x}_j})} + p_{\gamma_{\mathbf{x}_i} \backslash \gamma_{\mathbf{x}_j}} + p_{\gamma_{\mathbf{x}_j} \backslash \gamma_{\mathbf{x}_i}}$. Now, accounting for the subspaces overlapping with the observed subspace of $\mathbf{x}_k$, we get 
\begin{equation*}
\begin{aligned}
p(\mathbf{x}_i,\mathbf{x}_j) = p_{S \backslash (\gamma_{\mathbf{x}_i} \bigcup \gamma_{\mathbf{x}_j} \bigcup \gamma_{\mathbf{x}_k})} & + p_{\gamma_{\mathbf{x}_k} \backslash (\gamma_{\mathbf{x}_i} \bigcup \gamma_{\mathbf{x}_j})} + p_{\gamma_{\mathbf{x}_i} \backslash (\gamma_{\mathbf{x}_j} \bigcup \gamma_{\mathbf{x}_k})} \\
& + p_{(\gamma_{\mathbf{x}_i} \bigcap \gamma_{\mathbf{x}_k}) \backslash \gamma_{\mathbf{x}_j}} + p_{\gamma_{\mathbf{x}_j} \backslash (\gamma_{\mathbf{x}_i} \bigcup \gamma_{\mathbf{x}_k})} + p_{(\gamma_{\mathbf{x}_j} \bigcap \gamma_{\mathbf{x}_k}) \backslash \gamma_{\mathbf{x}_i}}.\\
\text{Similarly, } 
p(\mathbf{x}_j,\mathbf{x}_k) = p_{S \backslash (\gamma_{\mathbf{x}_i} \bigcup \gamma_{\mathbf{x}_j} \bigcup \gamma_{\mathbf{x}_k})} & + p_{\gamma_{\mathbf{x}_i} \backslash (\gamma_{\mathbf{x}_j} \bigcup \gamma_{\mathbf{x}_k})} + p_{\gamma_{\mathbf{x}_j} \backslash (\gamma_{\mathbf{x}_i} \bigcup \gamma_{\mathbf{x}_k})} \\
& + p_{(\gamma_{\mathbf{x}_i} \bigcap \gamma_{\mathbf{x}_j}) \backslash \gamma_{\mathbf{x}_k}} + p_{\gamma_{\mathbf{x}_k} \backslash (\gamma_{\mathbf{x}_i} \bigcup \gamma_{\mathbf{x}_j})} + p_{(\gamma_{\mathbf{x}_i} \bigcap \gamma_{\mathbf{x}_k}) \backslash \gamma_{\mathbf{x}_j}},\\
\text{and }
p(\mathbf{x}_k,\mathbf{x}_i) = p_{S \backslash (\gamma_{\mathbf{x}_i} \bigcup \gamma_{\mathbf{x}_j} \bigcup \gamma_{\mathbf{x}_k})} & + p_{\gamma_{\mathbf{x}_j} \backslash (\gamma_{\mathbf{x}_i} \bigcup \gamma_{\mathbf{x}_k})} + p_{\gamma_{\mathbf{x}_k} \backslash (\gamma_{\mathbf{x}_i} \bigcup \gamma_{\mathbf{x}_j})} \\
& + p_{(\gamma_{\mathbf{x}_j} \bigcap \gamma_{\mathbf{x}_k}) \backslash \gamma_{\mathbf{x}_i}} + p_{\gamma_{\mathbf{x}_i} \backslash (\gamma_{\mathbf{x}_j} \bigcup \gamma_{\mathbf{x}_k})} + p_{(\gamma_{\mathbf{x}_i} \bigcap \gamma_{\mathbf{x}_j}) \backslash \gamma_{\mathbf{x}_k}}.\\
\end{aligned}
\end{equation*}
Hence, after canceling out appropriate terms, we get
\begin{equation*}
\rho_{i,j,k} = p_{S \backslash (\gamma_{\mathbf{x}_i} \bigcup \gamma_{\mathbf{x}_j} \bigcup \gamma_{\mathbf{x}_k})} + p_{\gamma_{\mathbf{x}_k} \backslash (\gamma_{\mathbf{x}_i} \bigcup \gamma_{\mathbf{x}_j})} + 2 p_{(\gamma_{\mathbf{x}_i} \bigcap \gamma_{\mathbf{x}_k}) \backslash \gamma_{\mathbf{x}_j}} + p_{\gamma_{\mathbf{x}_i} \backslash (\gamma_{\mathbf{x}_j} \bigcup \gamma_{\mathbf{x}_k})} + p_{\gamma_{\mathbf{x}_j} \backslash (\gamma_{\mathbf{x}_i} \bigcup \gamma_{\mathbf{x}_k})}.
\end{equation*}
Now, since
\begin{equation*}
p_{\gamma_{\mathbf{x}_i} \backslash (\gamma_{\mathbf{x}_j} \bigcup \gamma_{\mathbf{x}_k})} + p_{\gamma_{\mathbf{x}_k} \backslash (\gamma_{\mathbf{x}_i} \bigcup \gamma_{\mathbf{x}_j})} + p_{(\gamma_{\mathbf{x}_i} \bigcap \gamma_{\mathbf{x}_k}) \backslash \gamma_{\mathbf{x}_j}} = p_{(\gamma_{\mathbf{x}_i} \bigcup \gamma_{\mathbf{x}_k}) \backslash \gamma_{\mathbf{x}_j}},
\end{equation*}
we can further simplify to
\begin{equation}\label{defRho}
\rho_{i,j,k} = p_{(\gamma_{\mathbf{x}_i} \bigcup \gamma_{\mathbf{x}_k})\backslash \gamma_{\mathbf{x}_j}} + p_{(\gamma_{\mathbf{x}_i} \bigcap \gamma_{\mathbf{x}_k})\backslash \gamma_{\mathbf{x}_j}} + p_{\gamma_{\mathbf{x}_j} \backslash (\gamma_{\mathbf{x}_i} \bigcup \gamma_{\mathbf{x}_k})} + p_{S \backslash (\gamma_{\mathbf{x}_i} \bigcup \gamma_{\mathbf{x}_j} \bigcup \gamma_{\mathbf{x}_k})}.
\end{equation}
Since all the terms in Expression (\ref{defRho}) must be either zero or positive, this proves that $\rho_{i,j,k} \geq 0$ $\forall$ $\mathbf{x}_i, \mathbf{x}_j, \mathbf{x}_k \in X$.
\end{proof}
\end{lemma}

\begin{lemma}\label{lemTri1}
For any three data points $\mathbf{x}_i, \mathbf{x}_j, \mathbf{x}_k \in X$, Inequality (\ref{eqnTri1}) is satisfied when $(\gamma_{\mathbf{x}_i} \bigcap \gamma_{\mathbf{x}_j})=(\gamma_{\mathbf{x}_j} \bigcap \gamma_{\mathbf{x}_k})=(\gamma_{\mathbf{x}_k} \bigcap \gamma_{\mathbf{x}_i})$.

\begin{proof}\label{pfLemTri1}
From Equation (\ref{eqnFwpd}) the Inequality (\ref{eqnTri1}) can be rewritten as
\begin{equation}\label{eqnTri2}
\begin{aligned}
(1-\alpha) \times \frac{d(\mathbf{x}_i,\mathbf{x}_j)}{d_{max}} + \alpha \times & p(\mathbf{x}_i,\mathbf{x}_j) \\
+ (1-\alpha) \times & \frac{d(\mathbf{x}_j,\mathbf{x}_k)}{d_{max}} + \alpha \times p(\mathbf{x}_j,\mathbf{x}_k) \\
\geq & (1-\alpha) \times \frac{d(\mathbf{x}_k,\mathbf{x}_i)}{d_{max}} + \alpha \times p(\mathbf{x}_k,\mathbf{x}_i).
\end{aligned}
\end{equation}
Further simplifying (\ref{eqnTri2}) by moving the penalty terms to the Left Hand Side (LHS) and the observed distance terms to the Right Hand Side (RHS), we get
\begin{equation}\label{eqnTri3}
\alpha \times \rho_{i,j,k} \geq \frac{(1-\alpha)}{d_{max}} \times (d(\mathbf{x}_k,\mathbf{x}_i) - (d(\mathbf{x}_i,\mathbf{x}_j)+d(\mathbf{x}_j,\mathbf{x}_k))).
\end{equation}
When $(\gamma_{\mathbf{x}_i} \bigcap \gamma_{\mathbf{x}_j})=(\gamma_{\mathbf{x}_j} \bigcap \gamma_{\mathbf{x}_k})=(\gamma_{\mathbf{x}_k} \bigcap \gamma_{\mathbf{x}_i})$, as $d(\mathbf{x}_i,\mathbf{x}_j)+d(\mathbf{x}_j,\mathbf{x}_k) \geq d(\mathbf{x}_k,\mathbf{x}_i)$, the RHS of Inequality (\ref{eqnTri3}) is less than or equal to zero. Now, it follows from Lemma \ref{penTri} that the LHS of Inequality (\ref{eqnTri3}) is always greater than or equal to zero as $\rho_{i,j,k} \geq 0$ and $\alpha \in (0,1]$. Hence, LHS $\geq$ RHS, which completes the proof.
\end{proof}
\end{lemma}

\begin{lemma}\label{lemTri2}
If $|\gamma_{\mathbf{x}_i} \bigcap \gamma_{\mathbf{x}_j}| \rightarrow 0$, $|\gamma_{\mathbf{x}_j} \bigcap \gamma_{\mathbf{x}_k}| \rightarrow 0$ and $|\gamma_{\mathbf{x}_k} \bigcap \gamma_{\mathbf{x}_i}| \rightarrow 0$, then Inequality (\ref{eqnTri3}) tends to be satisfied.

\begin{proof}\label{pfLemTri2}
When $|\gamma_{\mathbf{x}_i} \bigcap \gamma_{\mathbf{x}_j}| \rightarrow 0$, $|\gamma_{\mathbf{x}_j} \bigcap \gamma_{\mathbf{x}_k}| \rightarrow 0$ and $|\gamma_{\mathbf{x}_k} \bigcap \gamma_{\mathbf{x}_i}| \rightarrow 0$, then LHS $\rightarrow \alpha^{+}$ and RHS $\rightarrow 0$ for the Inequality (\ref{eqnTri3}). As $\alpha \in (0,1]$, Inequality (\ref{eqnTri3}) tends to be satisfied.
\end{proof}
\end{lemma}
The following lemma deals with the value of the parameter $\alpha \in (0,1]$ for which a relaxed form of the triangle inequality is satisfied for any three data instances in a dataset $X$.


\begin{lemma}\label{lemTri3}
Let $\mathcal{P} = min \{\rho_{i,j,k}:\mathbf{x}_i, \mathbf{x}_j, \mathbf{x}_k \in X, \rho_{i,j,k} > 0\}.$ Then, for any arbitrary constant $\epsilon$ satisfying $0 \leq \epsilon \leq \mathcal{P}$, if $\alpha \geq (1-\epsilon)$, then the following relaxed form of the triangle inequality
\begin{equation}\label{eqnRelTri1}
\delta(\mathbf{x}_i,\mathbf{x}_j) + \delta(\mathbf{x}_j,\mathbf{x}_k) \geq \delta(\mathbf{x}_k,\mathbf{x}_i) - {\epsilon}^2,
\end{equation}
is satisfied for any $\mathbf{x}_i, \mathbf{x}_j, \mathbf{x}_k \in X$.

\begin{proof}\label{pflemTri3}
\begin{enumerate}
\item If $\mathbf{x}_i$, $\mathbf{x}_j$, and $\mathbf{x}_k$ are all fully observed, then Inequality (\ref{eqnTri1}) holds. Now, since $\epsilon \geq 0$, therefore $\delta(\mathbf{x}_k,\mathbf{x}_i) \geq \delta(\mathbf{x}_k,\mathbf{x}_i) - {\epsilon}^2$. This implies $\delta(\mathbf{x}_i,\mathbf{x}_j) + \delta(\mathbf{x}_j,\mathbf{x}_k) \geq \delta(\mathbf{x}_k,\mathbf{x}_i) \geq \delta(\mathbf{x}_k,\mathbf{x}_i) - {\epsilon}^2$. Hence, Inequality (\ref{eqnRelTri1}) must hold.
\item If $(\gamma_{\mathbf{x}_i} \bigcap \gamma_{\mathbf{x}_j} \bigcap \gamma_{\mathbf{x}_k}) \neq S$ i.e. at least one of the data instances is not fully observed, and $\rho_{i,j,k} = 0$, then $(\gamma_{\mathbf{x}_i} \bigcup \gamma_{\mathbf{x}_k})\backslash \gamma_{\mathbf{x}_j} = \phi$, $(\gamma_{\mathbf{x}_i} \bigcap \gamma_{\mathbf{x}_k})\backslash \gamma_{\mathbf{x}_j} = \phi$, $S \backslash (\gamma_{\mathbf{x}_i} \bigcup \gamma_{\mathbf{x}_j} \bigcup \gamma_{\mathbf{x}_k}) = \phi$, and \\ $\gamma_{\mathbf{x}_j} \backslash (\gamma_{\mathbf{x}_i} \bigcup \gamma_{\mathbf{x}_k}) = \phi$. This implies that $\gamma_{\mathbf{x}_j} = S$, and $\gamma_{\mathbf{x}_k} \bigcup \gamma_{\mathbf{x}_i} = \gamma_{\mathbf{x}_j}$. Moreover, since $\rho_{i,j,k} = 0$, we have $\delta(\mathbf{x}_i,\mathbf{x}_j) + \delta(\mathbf{x}_j,\mathbf{x}_k) - \delta(\mathbf{x}_k,\mathbf{x}_i) = d(\mathbf{x}_i,\mathbf{x}_j) + d(\mathbf{x}_j,\mathbf{x}_k) - d(\mathbf{x}_k,\mathbf{x}_i)$. Now, $\gamma_{\mathbf{x}_i} \bigcap \gamma_{\mathbf{x}_k} \subseteq \gamma_{\mathbf{x}_i}$, $\gamma_{\mathbf{x}_i} \bigcap \gamma_{\mathbf{x}_k} \subseteq \gamma_{\mathbf{x}_k}$ and $\gamma_{\mathbf{x}_i} \bigcap \gamma_{\mathbf{x}_k} \subseteq \gamma_{\mathbf{x}_j}$ as $\gamma_{\mathbf{x}_k} \bigcup \gamma_{\mathbf{x}_i} = \gamma_{\mathbf{x}_j} =  S$. Therefore, $d(\mathbf{x}_i,\mathbf{x}_j) + d(\mathbf{x}_j,\mathbf{x}_k) - d(\mathbf{x}_k,\mathbf{x}_i) \geq d_{\gamma_{\mathbf{x}_i} \bigcap \gamma_{\mathbf{x}_k}}(\mathbf{x}_i,\mathbf{x}_j) + d_{\gamma_{\mathbf{x}_i} \bigcap \gamma_{\mathbf{x}_k}}(\mathbf{x}_j,\mathbf{x}_k) - d_{\gamma_{\mathbf{x}_i} \bigcap \gamma_{\mathbf{x}_k}}(\mathbf{x}_k,\mathbf{x}_i)$. Now, by the triangle inequality in subspace $\gamma_{\mathbf{x}_i} \bigcap \gamma_{\mathbf{x}_k}$, $d_{\gamma_{\mathbf{x}_i} \bigcap \gamma_{\mathbf{x}_k}}(\mathbf{x}_i,\mathbf{x}_j) + d_{\gamma_{\mathbf{x}_i} \bigcap \gamma_{\mathbf{x}_k}}(\mathbf{x}_j,\mathbf{x}_k) - d_{\gamma_{\mathbf{x}_i} \bigcap \gamma_{\mathbf{x}_k}}(\mathbf{x}_k,\mathbf{x}_i) \geq 0$. Hence, $\delta(\mathbf{x}_i,\mathbf{x}_j) + \delta(\mathbf{x}_j,\mathbf{x}_k) - \delta(\mathbf{x}_k,\mathbf{x}_i) \geq 0$, i.e. Inequalities (\ref{eqnTri1}) and (\ref{eqnRelTri1}) are satisfied.
\item If $(\gamma_{\mathbf{x}_i} \bigcap \gamma_{\mathbf{x}_j} \bigcap \gamma_{\mathbf{x}_k}) \neq S$ and $\rho_{i,j,k} \neq 0$, as $\alpha \geq (1-\epsilon)$, LHS of Inequality (\ref{eqnTri3}) $\geq (1-\epsilon) \times (p_{(\gamma_{\mathbf{x}_i} \bigcup \gamma_{\mathbf{x}_k})\backslash \gamma_{\mathbf{x}_j}} + p_{(\gamma_{\mathbf{x}_i} \bigcap \gamma_{\mathbf{x}_k})\backslash \gamma_{\mathbf{x}_j}} + p_{\gamma_{\mathbf{x}_j} \backslash (\gamma_{\mathbf{x}_i} \bigcup \gamma_{\mathbf{x}_k})} + p_{S \backslash (\gamma_{\mathbf{x}_i} \bigcup \gamma_{\mathbf{x}_j} \bigcup \gamma_{\mathbf{x}_k})})$. Since $\epsilon \leq \mathcal{P}$, we further get that LHS $\geq (1-\epsilon)\epsilon$. Moreover, as $\frac{1}{d_{max}}(d(\mathbf{x}_k,\mathbf{x}_i) - (d(\mathbf{x}_i,\mathbf{x}_j) + d(\mathbf{x}_j,\mathbf{x}_k))) \leq 1$, we get RHS of Inequality (\ref{eqnTri3}) $\leq \epsilon$. Therefore, LHS - RHS $\geq (1-\epsilon)\epsilon - \epsilon = -{\epsilon}^2$. Now, as Inequality (\ref{eqnTri3}) is obtained from Inequality (\ref{eqnTri1}) after some algebraic manipulation, it must hold that (LHS - RHS) of Inequality (\ref{eqnTri3}) = (LHS - RHS) of Inequality (\ref{eqnTri1}). Hence, we get
    $\delta(\mathbf{x}_i,\mathbf{x}_j) + \delta(\mathbf{x}_j,\mathbf{x}_k) - \delta(\mathbf{x}_k,\mathbf{x}_i) \geq -{\epsilon}^2$ which can be simplified to obtain Inequality (\ref{eqnRelTri1}). This completes the proof.
\end{enumerate}
\end{proof}
\end{lemma}

\par Let us now elucidate the proposed FWP (and consequently the proposed FWPD measure) by using the following example.
\begin{eg}
\end{eg}

Let $X \subset \mathbb{R}^3$ be a dataset consisting of $n=5$ data points, each having three features ($S=\{1,2,3\}$), some of which (marked by '*') are unobserved. The dataset is presented below (along with the feature observation counts and the observed feature sets for each of the instances).

\begin{small}
\begin{center}
\begin{tabular*}{0.551\textwidth}{|c|c c c|c|}
\hline
Data Point & $x_{i,1}$ & $x_{i,2}$ & $x_{i,3}$ & $\gamma_{\mathbf{x}_i}$ \\
\hline
$\mathbf{x}_1$ & * & $3$ & $2$ & $\{2,3\}$ \\
$\mathbf{x}_2$ & $1.2$ & * & $4$ & $\{1,3\}$ \\
$\mathbf{x}_3$ & * & $0$ & $0.5$ & $\{2,3\}$ \\
$\mathbf{x}_4$ & $2.1$ & $3$ & $1$ & $\{1,2,3\}$ \\
$\mathbf{x}_5$ & $-2$ & * & * & $\{1\}$ \\
\hline
Obs. Count & $w_1=3$ & $w_2=3$ & $w_3=4$ & - \\
\hline
\end{tabular*}
\end{center}
\end{small}
The pairwise observed distance matrix $A_{d}$ and the pairwise penalty matrix $A_{p}$, are as follows:
\begin{equation*}
A_{d} = \left[ \begin{matrix} 0, & 2, & 3.35, & 1, & 0 \\ 2, & 0, & 3.5, & 3.13, & 3.2 \\ 3.35, & 3.5, & 0, & 3.04, & 0 \\ 1, & 3.13, & 3.04, & 0, & 4.1 \\ 0, & 3.2, & 0, & 4.1, & 0 \end{matrix} \right] \text{ and } A_{p} = \left[ \begin{matrix} 0.3, & 0.6, & 0.3, & 0.3, & 1 \\ 0.6, & 0.3, & 0.6, & 0.3, & 0.7 \\ 0.3, & 0.6, & 0.3, & 0.3, & 1 \\ 0.3, & 0.3, & 0.3, & 0, & 0.7 \\ 1, & 0.7, & 1, & 0.7, & 0.7 \end{matrix} \right].
\end{equation*}
From $A_{d}$ it is observed that the maximum pairwise observed distance $d_{max}=4.1$. Then, the normalized observed distance matrix $A_{\bar{d}}$ is
\begin{equation*}
A_{\bar{d}} = \left[ \begin{matrix} 0, & 0.49, & 0.82, & 0.24, & 0 \\ 0.49, & 0, & 0.85, & 0.76, & 0.78 \\ 0.82, & 0.85, & 0, & 0.74, & 0 \\ 0.24, & 0.76, & 0.74, & 0, & 1 \\ 0, & 0.78, & 0, & 1, & 0 \end{matrix} \right].
\end{equation*}
$\mathcal{P} = 0.3$. While it is not necessary, let us choose $\alpha=0.7$. Using Equation (\ref{eqnFwpd}) to calculate the FWPD matrix $A_{\delta}$, we get:
\begin{equation*}
A_{\delta} = 0.3 \times A_{\bar{d}} + 0.7 \times A_{p} = \left[ \begin{matrix} 0.21, & 1.02, & 1.22, & 0.51, & 0.7 \\ 1.02, & 0.21, & 1.47, & 1.15, & 1.45 \\ 1.22, & 1.47, & 0.21, & 1.12, & 0.7 \\ 0.51, & 1.15, & 1.12, & 0, & 1.72 \\ 0.7, & 1.45, & 0.7, & 1.72, & 0.49 \end{matrix} \right].
\end{equation*}
It should be noted that in keeping with the properties of the FWPD described in Subsection \ref{sec:propFwpd}, $A_{\delta}$ is a symmetric matrix with the diagonal elements being the smallest entries in their corresponding rows (and columns) and the diagonal element corresponding to the fully observed point $\mathbf{x}_4$ being the only zero element. Moreover, it can be easily checked that the relaxed form of the triangle inequality, as given in Inequality (\ref{eqnRelTri1}), is always satisfied.

\section{k-means Clustering for Datasets with Missing Features using the proposed FWPD}\label{sec:Kmeans}

This section presents a reformulation of the k-means clustering problem 
for datasets with missing features, using the FWPD measure proposed in Section \ref{sec:fwpd}. The important notations used in this section (and beyond) are summarized in Table \ref{tabNota2}. The k-means problem (a term coined by MacQueen \citeyear{macqueen1967some}) deals with the partitioning of a set of $n$ data instances into $k(< n)$ clusters so as to minimize the sum of within-cluster dissimilarities. The standard heuristic algorithm to solve the k-means problem, referred to as the \emph{k-means algorithm}, was first proposed by Lloyd in 1957 \cite{lloyd1982least}, and rediscovered by Forgy \citeyear{forgy1965cluster}. Starting with $k$ random assignments of each of the data instances to one of the $k$ clusters, the k-means algorithm functions by iteratively recalculating the $k$ cluster centroids and reassigning the data instances to the nearest cluster (the cluster corresponding to the nearest cluster centroid), in an alternating manner. Selim \& Ismail \citeyear{selim1984kmeans} showed that the k-means algorithm converges to a local optimum of the non-convex optimization problem posed by the k-means problem, when the dissimilarity used is the Euclidean distance between data points.

\begin{table}[h]
\begin{center}
\caption{Some important notations used in Section \ref{sec:Kmeans} and beyond.}\label{tabNota2}
\begin{tabular}{c | c | l}
\hline
Notation & $\begin{array}{c c c}
\text{Counter-part in}\\
\text{k-means-FWPD}\\
\text{iteration } t\\
\end{array}$ & Meaning\\
\hline
$k$ & - & Number of clusters for k-means\\
$C_j$ & $C^t_j$ & $j$-th cluster for k-means\\
$u_{i,j}$ & $u^t_{i,j}$ & Membership of the data point $\mathbf{x}_i$ in the cluster $C_j$\\
$U$ & $U^t$ & $n \times k$ matrix of cluster memberships\\
$\mathcal{U}$ & - & Set of all possible $U$ values\\
$\mathbf{z}_j$ & $\mathbf{z}^t_j$ & Centroid of cluster $C_j$\\
$z_{j,l}$ & $z^t_{j,l}$ & $l$-th feature of the cluster centroid $\mathbf{z}_j$\\
$Z$ & $Z^t$ & Set of cluster centroids\\
$\mathcal{Z}$ & - & Set of all possible $Z$ values\\
$f(U,Z)$ & $f(U^t,Z^t)$ & k-means objective function defined on $\mathcal{U} \times \mathcal{Z}$\\
$X_l$ & - & Set of all $\mathbf{x}_i \in X$ having observed values for feature $l$\\
$U^*$ & - & Final cluster memberships found by k-means-FWPD\\
$Z^*$ & - & Final cluster centroids found by k-means-FWPD\\
$T$ & - & The convergent iteration of k-means-FWPD\\
- & $\tau$ & Any iteration preceding the current iteration $t$\\
$\mathcal{F}(Z)$ & - & Set of feasible membership matrices for $Z$\\
$\mathcal{F}(U)$ & - & Set of feasible centroid sets for $U$\\
$\mathcal{S}(U)$ & - & Set of super-feasible centroids sets for $U$\\
$(\tilde{U},\tilde{Z})$ & - & A partial optimal solution of the k-means-FWPD problem\\
$D$ & - & A feasible direction of movement for $U^*$\\
$\mathcal{O}$ & - & Big O notation\\
\hline
\end{tabular}
\end{center}
\end{table}

\par The proposed formulation of the k-means problem for datasets with missing features using the proposed FWPD measure, referred to as the \emph{k-means-FWPD problem} hereafter, differs from the standard k-means problem not only in that the underlying dissimilarity measure used is FWPD (instead of Euclidean distance), but also in the addition of a new constraint which ensures that a cluster centroid has observable values for exactly those features which are observed for at least one of the points in its corresponding cluster. Therefore, the k-means-FWPD problem to partition the dataset $X$ into $k$ clusters $(2 \leq k < n)$, can be formulated in the following way:
\begin{subequations}\label{eqnKmeansForm}
\begin{flalign}
\text{P: minimize } f(U,&Z)=\sum_{i=1}^{n} \sum_{j=1}^{k} u_{i,j}((1-\alpha)\times \frac{d(\mathbf{x}_{i},\mathbf{z}_{j})}{d_{max}} + \alpha \times p(\mathbf{x}_{i},\mathbf{z}_{j})), \\ 
\text{subject to } & \sum_{j=1}^{k} u_{i,j}=1 \text{ } \forall \text{ } i \in \{1, 2, \cdots, n\}, \label{constr1}\\ 
& u_{i,j} \in \{0,1\} \text{ } \forall \text{ } i \in \{1, 2, \cdots, n\},j \in \{1, 2, \cdots, k\}, \label{constr2}\\ 
\text{and }& \gamma_{\mathbf{z}_j} = \bigcup_{\mathbf{x}_{i} \in C_{j}} \gamma_{\mathbf{x}_i} \text{ } \forall \text{ } j \in \{1, 2, \cdots, k\}, \label{constr3} 
\end{flalign}
\end{subequations}
where $U=[u_{i,j}]$ is the $n \times k$ real matrix of memberships, $d_{max}$ denotes the maximum observed distance between any two data points $\mathbf{x}_i,\mathbf{x}_i \in X$, $\gamma_{\mathbf{z}_j}$ denotes the set of observed features for $\mathbf{z}_j$ $(j \in \{1, 2, \cdots, k\})$, $C_{j}$ denotes the $j$-th cluster (corresponding to the centroid $\mathbf{z}_j$), $Z=\{\mathbf{z}_1, \cdots, \mathbf{z}_k\}$, and it is said that $\mathbf{x}_i \in C_{j}$ when $u_{i,j}=1$.

\subsection{The k-means-FWPD Algorithm}\label{sec:KmeansFwpdAlgo}

To find a solution to the problem P, which is a non-convex program, we propose a Lloyd's heuristic-like algorithm based on the FWPD (referred to as \emph{k-means-FWPD algorithm}), as follows:
\begin{enumerate}
 \item Start with a random initial set of cluster assignments $U$ such that $\sum_{j=1}^{k} u_{i,j}=1$. Set $t=1$ and specify the maximum number of iterations $MaxIter$.
 \item For each cluster $C_{j}^{t}$ $(j = 1, 2, \cdots, k)$, calculate the observed features of the cluster centroid $\mathbf{z}_{j}^{t}$. The value for the $l$-th feature of a centroid $\mathbf{z}_{j}^{t}$ should be the average of the corresponding feature values for all the data instances in the cluster $C_{j}^{t}$ having observed values for the $l$-th feature. If none of the data instances in $C_{j}^{t}$ have observed values for the feature in question, then the value $z_{j,l}^{t-1}$ of the feature from the previous iteration should be retained. Therefore, the feature values are calculated as follows:
     \begin{equation}\label{eqnClustCentroid}
     z_{j,l}^{t}=\left\{ \begin{array}{cl}
     (\underset{\mathbf{x}_i \in X_l}{\sum}\; u_{i,j}^{t} \times x_{i,l})\bigg/(\underset{\mathbf{x}_i \in X_l}{\sum}\; u_{i,j}^{t}) \text{ }, & \mbox{$\forall \text{ } l \in \bigcup_{\mathbf{x}_i \in C_{j}^{t}} \gamma_{\mathbf{x}_i}$},\\
     \;\;\;\;\;\;\;\;\;z_{j,l}^{t-1}\;\;\;\;\;\;\;\;\;, & \mbox{$\forall \text{ } l \in \gamma_{\mathbf{z}_j^{t-1}} \backslash \bigcup_{\mathbf{x}_i \in C_{j}^{t}} \gamma_{\mathbf{x}_i}$},\\
     \end{array} \right.
     \end{equation}
     where $X_l$ denotes the set of all $\mathbf{x}_i \in X$ having observed values for the feature $l$.
 \item Assign each data point $\mathbf{x}_i$ $(i=1, 2, \cdots, n)$ to the cluster corresponding to its nearest (in terms of FWPD) centroid, i.e.
 	\begin{equation*}
     u_{i,j}^{t+1} =
     \left\{
     \begin{array}{ll}
     1, & \mbox{if $\mathbf{z}_{j}^{t}=\underset{\mathbf{z} \in Z^t}{\argmin}\; \delta(\mathbf{x}_i,\mathbf{z})$},\\
     0, & \mbox{otherwise}.\\
     \end{array} \right.
     \end{equation*}
     Set $t=t+1$. If $U^{t}=U^{t-1}$ or $t = MaxIter$, then go to Step 4; otherwise go to Step 2.
 \item Calculate the final cluster centroid set $Z^*$ as:
     \begin{equation}\label{eqnFinClustCentroid}
     z_{j,l}^{*}=\frac{\underset{\mathbf{x}_i \in X_l}{\sum}\; u_{i,j}^{t+1} \times       x_{i,l}}{\underset{\mathbf{x}_i \in X_l}{\sum}\; u_{i,j}^{t+1}} \text{ } \forall \text{ } l \in \bigcup_{\mathbf{x}_i \in C_{j}^{t+1}} \gamma_{\mathbf{x}_i}.
     \end{equation}
     Set $U^* = U^{t+1}$.
\end{enumerate}

\begin{figure}[!ht]
\centering
\subfloat[Traditional k-means.]{\includegraphics[width=2.25 in]{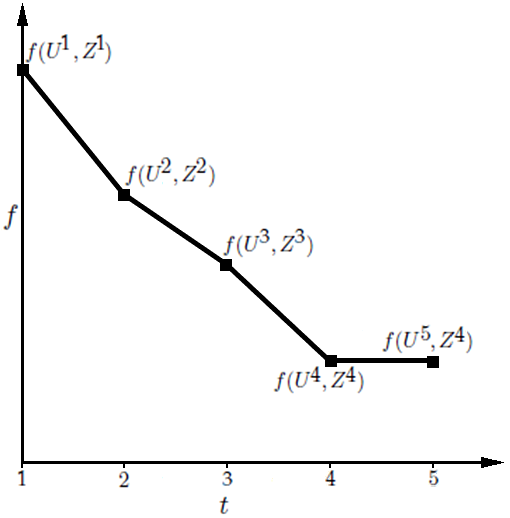}
\label{figConvTrad}}
\hfil
\subfloat[k-means-FWPD.]{\includegraphics[width=2.25 in]{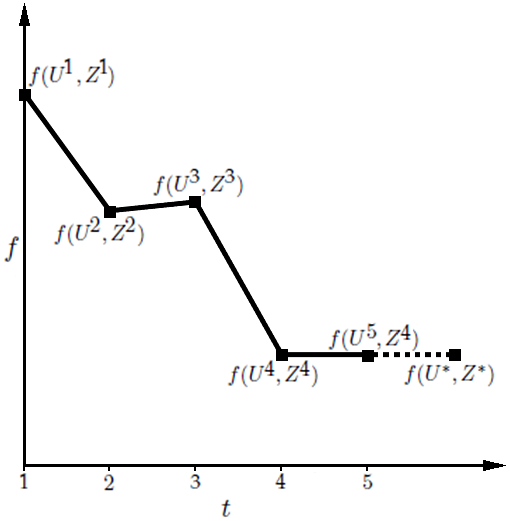}
\label{figConvNew}}
\caption{Comparison of convergences of traditional k-means and k-means-FWPD algorithms.}
\label{figConv}
\end{figure}

\begin{rmk}
The iterations of the traditional k-means algorithm are known to each result in a decrease in the value of the objective function $f$ \cite{selim1984kmeans} (Figure \ref{figConvTrad}). However, for the k-means-FWPD algorithm, the $Z^t$ calculations for some of the iterations may result in a finite increase in $f$, as shown in Figure \ref{figConvNew}. We show in Theorem \ref{thmFinFin} that only a finite number of such increments may occur during a given run of the algorithm, thus ushering in ultimate convergence. Moreover, the final feasible, locally-optimal solution is obtained using Step 4 (denoted by dotted line) which does not result in any further change to the objective function value.
\end{rmk}

\subsection{Notions of Feasibility in Problem P}

Let $\mathcal{U}$ and $\mathcal{Z}$ respectively denote the sets of all possible $U$ and $Z$. Unlike the traditional k-means problem, the entire $\mathcal{U} \times \mathcal{Z}$ space is not feasible for the Problem P. There exists a set of feasible $U$ for a given $Z$. Similarly, there exist sets of feasible and super-feasible $Z$ (a super-set of the set of feasible $Z$) for a given $U$. In this subsection, we formally define these notions.

\begin{dfn}\label{dfnFeasU}
Given a cluster centroid set $Z$, the set $\mathcal{F}(Z)$ of feasible membership matrices is given by
\begin{equation*}
\mathcal{F}(\hat{Z}) = \{U : u_{i,j} = 0 \text{ } \forall j \in \{1,2, \cdots, k\} \text{ such that } \gamma_{\mathbf{z}_j} \subset \gamma_{\mathbf{x}_i} \},
\end{equation*}
i.e. $\mathcal{F}(Z)$ is the set of all such membership matrices which do not assign any $\mathbf{x}_i \in X$ to a centroid in $Z$ missing some feature $l \in \gamma_{\mathbf{x}_i}$.
\end{dfn}

\begin{dfn}\label{dfnFeasZ}
Given a membership matrix $U$, the set $\mathcal{F}(U)$ of feasible cluster centroid sets can be defined as
\begin{equation*}
\mathcal{F}(U) = \{Z : Z \text{ satisfies constraint } (\ref{constr3}) \}.
\end{equation*}
\end{dfn}

\begin{dfn}\label{dfnSupfeasZ}
Given a membership matrix $U$, the set $\mathcal{S}(U)$ of super-feasible cluster centroid sets is
\begin{equation}\label{constr3rel}
\mathcal{S}(U) = \{Z : \gamma_{\mathbf{z}_j} \supseteq \bigcup_{\mathbf{x}_{i} \in C_{j}} \gamma_{\mathbf{x}_i} \text{ } \forall \text{ } j \in \{1, 2, \cdots, k\} \},
\end{equation}
i.e. $\mathcal{S}(U)$ is the set of all such centroid sets which ensure that any centroid has observed values at least for those features which are observed for any of the points assigned to its corresponding cluster in $U$.
\end{dfn}

\begin{figure}[!t]
\centering
\includegraphics[width=0.725\textwidth]{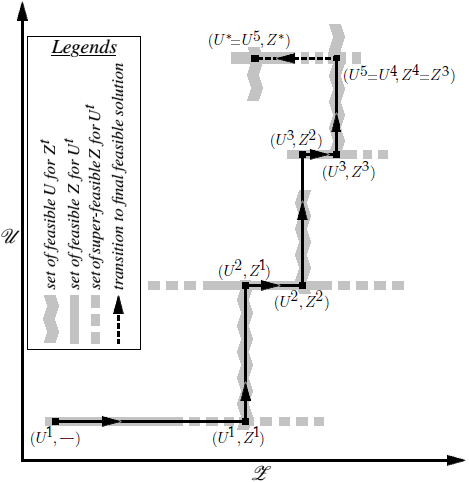}
\caption{Simplified representation of how the k-means-FWPD algorithm traverses the $\mathcal{U} \times \mathcal{Z}$ space ($\mathcal{U}$ and $\mathcal{Z}$ are shown to be unidimensional for the sake of visualizability}).
\label{fig:algo1}
\end{figure}

\begin{rmk}
The k-means-FWPD problem differs from traditional k-means in that not all $U \in \mathcal{U}$ are feasible for a given $Z$. Additionally, for a given $U$, there exists a set $\mathcal{S}(U)$ of super-feasible $Z$; $\mathcal{F}(U)$ a subset of $\mathcal{S}(U)$ being the set of feasible $Z$. The traversal of the k-means-FWPD algorithm is illustrated in Figure \ref{fig:algo1} where the grey solid straight lines denote the set of feasible $Z$ for the current $U^t$ while the rest of the super-feasible region is marked by the corresponding grey dotted straight line. Furthermore, the grey jagged lines denote the feasible set of $U$ for the current $Z^t$. Starting with a random $U^1 \in \mathcal{U}$ (Step 1), the algorithm finds $Z^1 \in \mathcal{S}(U^1)$ (Step 2), $U^2 \in \mathcal{F}(Z^1)$ (Step 3), and $Z^2 \in \mathcal{F}(U^2)$ (Step 2). However, it subsequently finds $U^3 \not\in \mathcal{F}(Z^2)$ (Step 3), necessitating a feasibility adjustment (see Section \ref{sec:feas}) while calculating $Z^3$ (Step 2). Subsequently, the algorithm converges to $(U^5,Z^4)$. For the convergent $(U^{T+1},Z^T)$, $U^{T+1} \in \mathcal{F}(Z^T)$ but it is possible that $Z^T \in \mathcal{S}(U^{T+1})\backslash \mathcal{F}(U^T)$ (as seen in the case of Figure \ref{fig:algo1}). However, the final $(U^*,Z^*)$ (obtained by the dotted black line transition denoting Step 4) is seen to be feasible in both respects and is shown (in Theorem \ref{thmLocalOpt}) to be locally-optimal in the corresponding feasible region.
\end{rmk}

\subsection{Partial Optimal Solutions}\label{sec:partOptSoln}

This subsection deals with the concept of \emph{partial optimal solutions} of the problem P, to one of which the k-means-FWPD algorithm is shown to converge (prior to Step 4). The following definition formally presents the concept of a partial optimal solution.

\begin{dfn}\label{dfnPOS}
A partial optimal solution $(\tilde{U},\tilde{Z})$ of problem P, satisfies the following conditions \cite{wendel1976minimization}:
\begin{equation*}
\begin{aligned}
& f(\tilde{U},\tilde{Z}) \leq f(U,\tilde{Z}) \text{ } \forall \text{ } U \in \mathcal{F}(\tilde{Z}) \text{ where } \tilde{U} \in \mathcal{F}(\tilde{Z}),\\
\text{and } & f(\tilde{U},\tilde{Z}) \leq f(\tilde{U},Z) \text{ } \forall \text{ } Z \in \mathcal{S}(\tilde{U}) \text{ where } \tilde{Z} \in \mathcal{S}(\tilde{U}).
\end{aligned}
\end{equation*}
\end{dfn}

To obtain a partial optimal solution of P, the two following subproblems are defined:
\begin{equation*}
\begin{aligned}
&\text{P1: Given } U \in \mathcal{U}, \text{ minimize } f(U,Z) \text{ over } Z \in \mathcal{S}(U).\\
&\text{P2: Given } Z \text{ satisfying (\ref{constr3}), minimize } f(U,Z) \text{ over } U \in \mathcal{U}.
\end{aligned}
\end{equation*}
The following lemmas establish that Steps 2 and 3 of the k-means-FWPD algorithm respectively solve the problems P1 and P2 for a given iterate. The subsequent theorem shows that the k-means-FWPD algorithm converges to a partial optimal solution of P.

%

\begin{lemma}\label{lemSolveP1}
Given a $U^t$, the centroid matrix $Z^t$ calculated using Equation (\ref{eqnClustCentroid}) is an optimal solution of the Problem P1.\\


\begin{proof}
For a fixed $U^t \in \mathcal{U}$, the objective function is minimized when $\frac{\partial f}{\partial z_{j,l}^t}=0 \text{ } \forall j \in \{1, \cdots, k\}, l \in \gamma_{\mathbf{z}^t_j}$. For a particular $\mathbf{z}^t_j$, it follows from Definition \ref{defFwp} that $\{p(\mathbf{x}_{i},\mathbf{z}^t_{j}): \mathbf{x}_i \in C^t_j\}$ is independent of the values of the features of $\mathbf{z}^t_j$, as $\gamma_{\mathbf{x}_i} \bigcap \gamma_{\mathbf{z}^t_j} = \gamma_{\mathbf{x}_i}\ \text{ } \forall \mathbf{x}_i \in C^t_j$. Since an observed feature $l \in \gamma_{\mathbf{z}^t_j} \backslash (\bigcup_{\mathbf{x}_{i} \in C^t_{j}} \gamma_{\mathbf{x}_i})$ of $\mathbf{z}^t_j$ has no contribution to the observed distances, $\frac{\partial f}{\partial z_{j,l}^t}=0 \text{ } \forall l \in \gamma_{\mathbf{z}^t_j} \backslash (\bigcup_{\mathbf{x}_{i} \in C^t_{j}} \gamma_{\mathbf{x}_i})$. For an observed feature $l\in \bigcup_{\mathbf{x}_{i} \in C^t_{j}} \gamma_{\mathbf{x}_i}$ of $\mathbf{z}^t_j$, differentiating $f(U^t,Z^t)$ w.r.t. $z_{j,l}^t$ we get
\begin{equation*}
\frac{\partial f}{\partial z_{j,l}^t} = \frac{(1-\alpha)}{d_{max}} \times \underset{\mathbf{x}_i \in X_l}{\sum} u_{i,j}^t(\frac{x_{i,l}-z_{j,l}^t}{d(\mathbf{x}_i,\mathbf{z}^t_j)}).
\end{equation*}
Setting $\frac{\partial f}{\partial z_{j,l}^t} = 0$ and solving for $z_{j,l}^t$, we obtain
\begin{equation*}
z_{j,l}^t=\frac{\underset{\mathbf{x}_i \in X_l}{\sum} u_{i,j}^t \times x_{i,l}}{\underset{\mathbf{x}_i \in X_l}{\sum} u_{i,j}^t}.
\end{equation*}
Since Equation (\ref{eqnClustCentroid}) is in keeping with this criterion and ensures that constraint (\ref{constr3rel}) is satisfied, the centroid matrix $Z^t$ calculated using Equation (\ref{eqnClustCentroid}) is an optimal solution of P1.
\end{proof}
\end{lemma}

\begin{lemma}\label{lemSolveP2}
For a given $Z^t$, problem P2 is solved if $u^{t+1}_{i,j}=1$ and $u^{t+1}_{i,j^{'}}=0$ $\forall$ $i \in \{1, \cdots, n\}$ when $\delta(\mathbf{x}_i,\mathbf{z}^t_j) \leq \delta(\mathbf{x}_i,\mathbf{z}^t_{j^{'}})$, for all $j^{'} \neq j$.\\

\begin{proof}
It is clear that the contribution of $\mathbf{x}_i$ to the total objective function is $\delta(\mathbf{x}_i,\mathbf{z}^t_j)$ when $u^{t+1}_{i,j}=1$ and $u^{t+1}_{i,j^{'}}=0$ $\forall$ $j^{'} \neq j$. Since any alternative solution is an extreme point of $\mathcal{U}$ \cite{selim1984kmeans}, it must satisfy (\ref{constr2}). Therefore, the contribution of $\mathbf{x}_i$ to the objective function for an alternative solution will be some $\delta(\mathbf{x}_i,\mathbf{z}^t_{j^{'}}) \geq \delta(\mathbf{x}_i,\mathbf{z}^t_j)$. Hence, the contribution of $\mathbf{x}_i$ is minimized by assigning $u^{t+1}_{i,j}=1$ and $u^{t+1}_{i,j^{'}}=0$ $\forall$ $j^{'} \neq j$. This argument holds true for all $\mathbf{x}_i \in X$, i.e. $\forall$ $i \in \{1, \cdots, n\}$. This completes the proof.
\end{proof}
\end{lemma}

\begin{theorem}\label{thmConvergePOS}
The k-means-FWPD algorithm finds a partial optimal solution of P.\\

\begin{proof}
Let $T$ denote the terminal iteration. Since Step 2 and Step 3 of the k-means-FWPD algorithm respectively solve P1 and P2, the algorithm terminates only when the obtained iterate $(U^{T+1},Z^{T})$ solves both P1 and P2. Therefore, $f(U^{T+1},Z^{T}) \leq f(U^{T+1},Z) \text{ } \forall Z \in \mathcal{S}(U^{T+1})$. Since Step 2 ensures that $Z^{T} \in \mathcal{S}(U^{T})$ and $U^{T+1} = U^{T}$, we must have $Z^{T} \in \mathcal{S}(U^{T+1})$. Moreover, $f(U^{T+1},Z^{T}) \leq f(U,Z^T)$ $\forall U \in \mathcal{U}$ which implies $f(U^{T+1},Z^{T}) \leq f(U,Z^T) \text{ } \forall U \in \mathcal{F}(Z^T)$. Now, Step 2 ensures that $\gamma_{\mathbf{z}^{T}_j} \supseteq \bigcup_{\mathbf{x}_{i} \in C^T_{j}} \gamma_{\mathbf{x}_i} \text{ } \forall \text{ } j \in \{1, 2, \cdots, k\}$. Since we must have $U^{T+1} = U^{T}$ for convergence to occur, it follows that $\gamma_{\mathbf{z}^{T}_j} \supseteq \bigcup_{\mathbf{x}_{i} \in C^{T+1}_{j}} \gamma_{\mathbf{x}_i} \text{ } \forall \text{ } j \in \{1, 2, \cdots, k\}$, hence $u^{T+1}_{i,j} = 1$ implies $\gamma_{\mathbf{z}^T_j} \supseteq \gamma_{\mathbf{x}_i}$. Therefore, $U^{T+1} \in \mathcal{F}(Z^T)$. Consequently, the terminal iterate of Step 3 of the k-means-FWPD algorithm must be a partial optimal solution of P.
\end{proof}
\end{theorem}

\subsection{Feasibility Adjustments}\label{sec:feas}

Since it is possible for the number of observed features of the cluster centroids to increase over the iterations to maintain feasibility w.r.t. constraint (\ref{constr3}), we now introduce the concept of \emph{feasibility adjustment}, the consequences of which are discussed in this subsection.

\begin{dfn}\label{dfnFA}
A feasibility adjustment for cluster $j$ ($j \in \{1,2,\cdots,k\}$) is said to occur in iteration $t$ if $\gamma_{\mathbf{z}^t_j} \supset \gamma_{\mathbf{z}^{t-1}_j}$ or $\gamma_{\mathbf{z}^t_j} \backslash \gamma_{\mathbf{z}^{t-1}_j} \neq \phi$, i.e. if the centroid $\mathbf{z}^t_j$ acquires an observed value for at least one feature which was unobserved for its counter-part $\mathbf{z}^{t-1}_j$ in the previous iteration.
\end{dfn}
The following lemma shows that feasibility adjustment can only occur for a cluster as a result of the addition of a new data point previously unassigned to it.

\begin{lemma}\label{lemNewPoint}
Feasibility adjustment occurs for a cluster $C_j$ in iteration $t$ iff at least one data point $\mathbf{x}_i$, such that $\gamma_{\mathbf{x}_i} \backslash \gamma_{\mathbf{z}^{\tau}_j} \neq \phi$ $\forall \tau < t$, which was previously unassigned to $C_j$ (i.e. $u^{\tau}_{i,j} = 0$ $\forall \tau < t$) is assigned to it in iteration $t$.\\

\begin{proof}
Due to Equation (\ref{eqnClustCentroid}), all features defined for $\mathbf{z}^{t-1}_j$ are also retained for $\mathbf{z}^t_j$. Therefore, for $\gamma_{\mathbf{z}^t_j} \backslash \gamma_{\mathbf{z}^{t-1}_j} \neq \phi$ there must exist some $\mathbf{x}_i$ such that $u^t_{i,j} = 1$, $u^{t-1}_{i,j} = 0$, and $\gamma_{\mathbf{x}_i} \backslash \gamma_{\mathbf{z}^{t-1}_j} \neq \phi$. Since the set of defined features for any cluster centroid is a monotonically growing set, we have $\gamma_{\mathbf{x}_i} \backslash \gamma_{\mathbf{z}^{\tau}_j} \neq \phi$ $\forall \tau < t$. It then follows from constraint (\ref{constr3}) that $u^{\tau}_{i,j} = 0$ $\forall \tau < t$. Now, to prove the converse, let us assume the existence of some $\mathbf{x}_i$ such that $\gamma_{\mathbf{x}_i} \backslash \gamma_{\mathbf{z}^{\tau}_j} \neq \phi$ $\forall \tau < t$ and $u^{\tau}_{i,j} = 0$ $\forall \tau < t$. Since $\gamma_{\mathbf{x}_i} \backslash \gamma_{\mathbf{z}^{t-1}_j} \neq \phi$ and $\gamma_{\mathbf{z}^t_j} \supseteq \gamma_{\mathbf{x}_i} \bigcup \gamma_{\mathbf{z}^{t-1}_j}$, it follows that $\gamma_{\mathbf{z}^t_j} \backslash \gamma_{\mathbf{z}^{t-1}_j} \neq \phi$.
\end{proof}
\end{lemma}
The following theorem deals with the consequences of the feasibility adjustment phenomenon.

\begin{theorem}\label{thmFinFin}
For a finite number of iterations during a single run of the k-means-FWPD algorithm, there may be a finite increment in the objective function $f$, due to the occurrence of feasibility adjustments.\\

\begin{proof}

It follows from Lemma \ref{lemSolveP1} that $f(U^t,Z^t) \leq f(U^t,Z)$ $\forall Z \in \mathcal{S}(U^t)$. If there is no feasibility adjustment in iteration $t$, $\mathcal{S}(U^{t-1}) = \mathcal{S}(U^t)$. Hence, $f(U^t,Z^t) \leq f(U^t,Z^{t-1})$. However, if a feasibility adjustment occurs in iteration $t$, then $\gamma_{\mathbf{z}^t_j} \subset \gamma_{\mathbf{z}^{t-1}_j}$ for at least one $j \in \{1,2,\cdots,k\}$. Hence, $Z^{t-1} \in \mathcal{Z} \backslash \mathcal{S}(U^t)$ and we may have $f(U^t,Z^t) > f(U^t,Z^{t-1})$. Since both $f(U^t,Z^t)$ and $f(U^t,Z^{t-1})$ are finite, $(f(U^t,Z^t) - f(U^t,Z^{t-1}))$ must also be finite. Now, the maximum number of feasibility adjustments occur in the worst case scenario where each data point, having an unique set of observed features (which are unobserved for all other data points), traverses all the clusters before convergence. Therefore, the maximum number of possible feasibility adjustments during a single run of the k-means-FWPD algorithm is $n(k-1)$, which is finite.
\end{proof}
\end{theorem}

\subsection{Convergence of the k-means-FWPD Algorithm}\label{sec:KmeansFwpdConverge}

We now show that the k-means-FWPD algorithm converges to the partial optimal solution, within a finite number of iterations. The following lemma and the subsequent theorem are concerned with this.



\begin{lemma}\label{lemConvOrFeas}
Starting with a given iterate $(U^t,Z^t)$, the k-means-FWPD algorithm either reaches convergence or encounters a feasibility adjustment, within a finite number of iterations.\\

\begin{proof}
Let us first note that there are a finite number of extreme points of $\mathcal{U}$. Then, we observe that an extreme point of $\mathcal{U}$ is visited at most once by the algorithm before either convergence or the next feasibility adjustment. Suppose, this is not true, and let $U^{t_1}=U^{t_2}$ for distinct iterations $t_1$ and $t_2$ $(t_1 \geq t, t_1 < t_2)$ of the algorithm. Applying Step 2 of the algorithm, we get $Z^{t_1}$ and $Z^{t_2}$ as optimal centroid sets for $U^{t_1}$ and $U^{t_2}$, respectively. Then, $f(U^{t_1},Z^{t_1}) = f(U^{t_2},Z^{t_2})$ since $U^{t_1}=U^{t_2}$. However, it is clear from Lemmas \ref{lemSolveP1}, \ref{lemSolveP2} and Theorem \ref{thmFinFin} that $f$ strictly decreases subsequent to the iterate $(U^t,Z^t)$ and prior to either the next feasibility adjustment (in which case the value of $f$ may increase) or convergence (in which case $f$ remains unchanged). Hence, $U^{t_1} \neq U^{t_2}$. Therefore, it is clear from the above argument that the k-means-FWPD algorithm either converges or encounters a feasibility adjustment within a finite number of iterations.
\end{proof}
\end{lemma}



\begin{theorem}\label{thmConvFwpd}
The k-means-FWPD algorithm converges to a partial optimal solution within a finite number of iterations.\\

\begin{proof}
It follows from Lemma \ref{lemConvOrFeas} that the first feasibility adjustment is encountered within a finite number of iterations since initialization and that each subsequent feasibility adjustment occurs within a finite number of iterations of the previous. Moreover, we know from Theorem \ref{thmFinFin} that there can only be a finite number of feasibility adjustments during a single run of the algorithm. Therefore, the final feasibility adjustment must occur within a finite number of iterations. Moreover, it follows from Lemma \ref{lemConvOrFeas} that the algorithm converges within a finite number of subsequent iterations. Hence, the k-means-FWPD algorithm must converge within a finite number of iterations.
\end{proof}
\end{theorem}

\subsection{Local Optimality of the Final Solution}\label{sec:Finoptim}

In this subsection, we establish the local optimality of the final solution obtained in Step 4 of the k-means-FWPD algorithm, subsequent to convergence in Step 3.

\begin{lemma}\label{lemZStaropt}
$Z^*$ is the unique optimal feasible cluster centroid set for $U^*$, i.e. $Z^* \in \mathcal{F}(U^*)$ and $f(U^*,Z^*) \leq f(U^*,Z) \text{ } \forall Z \in \mathcal{F}(U^*)$.

\begin{proof}
Since $Z^*$ satisfies constraint (\ref{constr3}) for $U^*$, $Z^* \in \mathcal{F}(U^*)$. We know from Lemma \ref{lemSolveP1} that for $f(U^*,Z^*) \leq f(U^*,Z) \text{ } \forall Z \in \mathcal{F}(U^*)$, we must have
\begin{equation*}
z^*_{j,l}=\frac{\underset{\mathbf{x}_i \in X_l}{\sum} u^*_{i,j} \times x_{i,l}}{\underset{\mathbf{x}_i \in X_l}{\sum} u^*_{i,j}}.
\end{equation*}
As this is ensured by Step 4, $Z^*$ must be the unique optimal feasible cluster centroid set for $U^*$.
\end{proof}
\end{lemma}

\begin{lemma}\label{lemFeasZSupfeasZ}
If $Z^*$ is the unique optimal feasible cluster centroid set for $U^*$, then $f(U^*,Z^*) \leq f(U,Z^*) \text{ } \forall U \in \mathcal{F}(Z^*)$.

\begin{proof}
We know from Theorem \ref{thmConvergePOS} that $f(U^{*},Z^{T}) \leq f(U,Z^T) \text{ } \forall U \in \mathcal{F}(Z^T)$. Now, $\gamma_{\mathbf{z}^*_j} \subseteq \gamma_{\mathbf{z}^T_j} \text{ } \forall j \in \{1,2,\cdots,k\}$. Therefore, $\mathcal{F}(Z^*) \subseteq \mathcal{F}(Z^T)$ must hold. It therefore follows that $f(U^*,Z^*) \leq f(U,Z^*) \text{ } \forall U \in \mathcal{F}(Z^*)$.
\end{proof}
\end{lemma}
Now, the following theorem shows that the final solution obtained by Step 4 of the k-means-FWPD algorithm is locally optimal.

\begin{theorem}\label{thmLocalOpt}
The final solution $(U^*,Z^*)$ obtained by Step 4 of the k-means-FWPD algorithm is a local optimal solution of P.

\begin{proof}
We have from Lemma \ref{lemFeasZSupfeasZ} that $f(U^*,Z^*) \leq f(U,Z^*) \text{ } \forall U \in \mathcal{F}(Z^*)$. Therefore, $f(U^*,Z^*) \leq \min_{U} \{f(U,Z^*):U \in \mathcal{F}(Z^*)\}$ which implies that for all feasible directions $D$ at $U^*$, the one-sided directional derivative \cite{lasdon2013optimization},
\begin{equation}\label{eqnTraceGeqZero}
trace(\nabla_{U}f(U^*,Z)^{\mathrm{T}}D) \geq 0.
\end{equation}
Now, since $Z^*$ is the unique optimal feasible cluster centroid set for $U^*$ (Lemma \ref{lemZStaropt}), $(U^*,Z^*)$ is a local optimum of problem P.
\end{proof}
\end{theorem}

\subsection{Time Complexity of the k-means-FWPD Algorithm}\label{sec:TimeCompKmeansFwpd}

In this subsection, we present a brief discussion on the time complexity of the k-means-FWPD algorithm. The k-means-FWPD algorithm consists of four basic steps, which are repeated iteratively. These steps are
\begin{enumerate}
 \item \emph{Centroid Calculation}: As a maximum of $m$ features of each centroid must be calculated, the complexity of centroid calculation is at most $\mathcal{O}(kmn)$.
 \item \emph{Distance Calculation}: As each distance calculation involves at most $m$ features, the observed distance calculation between $n$ data instances and $k$ cluster centroids is at most $\mathcal{O}(kmn)$.
 \item \emph{Penalty Calculation}: The penalty calculation between a data point and a cluster centroid involves at most $m$ summations. Hence, penalty calculation over all possible pairings is at most $\mathcal{O}(kmn)$.
 \item \emph{Cluster Assignment}: The assignment of $n$ data points to $k$ clusters consists of the comparisons of the dissimilarities of each point with $k$ clusters, which is $\mathcal{O}(nk)$.
\end{enumerate}
Therefore, if the algorithm runs for $T$ iterations, the total computational complexity is $\mathcal{O}(kmnT)$ which is the same as that of the standard k-means algorithm.

\section{Hierarchical Agglomerative Clustering for Datasets with Missing Features using the proposed FWPD}\label{sec:Hier}

In this section we present HAC clustering methods using the proposed FWPD measure that can be directly applied to data with missing features. The important notations used in this section (and beyond) are summarized in Table \ref{tabNota3}.

\begin{table}[h]
\begin{center}
\caption{Some important notations used in Section \ref{sec:Hier} and beyond.}\label{tabNota3}
\begin{tabular}{c | l}
\hline
Notation & Meaning\\
\hline
$B^t$ & Set of hierarchical clusters obtained in iteration $t$ of HAC-FWPD\\
$\beta^t_i$ & $i$-th hierarchical cluster in $B^t$\\
$Q^t$ & Matrix of dissimilarities between the hierarchical clusters in $B^t$\\
$q^(i,j)$ & $(i,j)$-th element of $Q^t$\\
$q^t_{min}$ & Smallest non-zero value in $Q^t$\\
$M$ & List of location in $Q^t$ having value $q^t_{min}$\\
$G$ & New hierarchical cluster formed by merging two of the closest hierarchical clusters in $B^t$\\
$i_G$ & Location of $G$ in the set $B^{t+1}$\\
$L(G,\beta)$ & Linkage between two hierarchical clusters $G$ and $\beta$\\
\hline
\end{tabular}
\end{center}
\end{table}
 
Hierarchical agglomerative schemes for data clustering seek to build a multi-level hierarchy of clusters, starting with each data point as a single cluster, by combining two (or more) of the most proximal clusters at one level to obtain a lower number of clusters at the next (higher) level. A survey of HAC methods can be found in \cite{murtagh2012algorithms}. However, these methods cannot be directly applied to datasets with missing features. Therefore, in this section, we develop variants of HAC methods, based on the proposed FWPD measure. Various proximity measures may be used to merge the clusters in an agglomerative clustering method. Modifications of the three most popular of such proximity measures (Single Linkage (SL), Complete Linkage (CL) and Average Linkage (AL)) so as to have FWPD as the underlying dissimilarity measure, are as follows:
\begin{enumerate}
 \item \emph{Single Linkage with FWPD (SL-FWPD)}: The SL between two clusters $C_i$ and $C_j$ is $\min \{\delta(\mathbf{x}_i,\mathbf{x}_j):\mathbf{x}_i \in C_i, \mathbf{x}_j \in C_j\}$.
 \item \emph{Complete Linkage with FWPD (CL-FWPD)}: The CL between two clusters $C_i$ and $C_j$ is $\max \{\delta(\mathbf{x}_i,\mathbf{x}_j):\mathbf{x}_i \in C_i, \mathbf{x}_j \in C_j\}$.
 \item \emph{Average Linkage with FWPD (AL-FWPD)}: $\frac{1}{|C_i| \times |C_j|} \underset{\mathbf{x}_i \in C_i}{\sum} \underset{\mathbf{x}_j \in C_j}{\sum} \delta(\mathbf{x}_i,\mathbf{x}_j)$ is the AL between two clusters $C_i$ and $C_j$, where $|C_i|$ and $|C_j|$ are respectively the number of instances in the clusters $C_i$ and $C_j$.
\end{enumerate}

\subsection{The HAC-FWPD Algorithm}\label{sec:AggFwpdAlgo}

To achieve hierarchical clusterings in the presence of unstructured missingness, the HAC method based on SL-FWPD, CL-FWPD, or AL-FWPD (referred to as the \emph{HAC-FWPD algorithm} hereafter) is as follows:
\begin{enumerate}
 \item Set $B^0 = X$. Compute pairwise dissimilarities $\delta(\mathbf{x}_i,\mathbf{x}_j)$, $\forall$ $\mathbf{x}_i,\mathbf{x}_j \in X$ and construct the dissimilarity matrix $Q^0$ so that $q^0(i,j) = \delta(\mathbf{x}_i,\mathbf{x}_j)$. Set $t=0$.
 \item Search $Q^t$ to identify the set $M = \{(i_1,j_1), (i_2,j_2), \cdots, (i_k,j_k)\}$ containing all the pairs of indexes such that $q^t(i_r,j_r) = q^t_{min}$ $\forall$ $r \in \{1, 2, \cdots, k\}$, $q^t_{min}$ being the smallest non-zero element in $Q^t$.
 \item Merge the elements corresponding to any one pair in $M$, say $\beta^t_{ir}$ and $\beta^t_{jr}$ corresponding to the pair $(i_r,j_r)$, into a single group $G = \{\beta^t_{ir}, \beta^t_{jr}\}$. Construct $B^{t+1}$ by removing $\beta^t_{ir}$ and $\beta^t_{jr}$ from $B^t$ and inserting $G$. 
 \item Define $Q^{t+1}$ on $B^{t+1} \times B^{t+1}$ as $q^{t+1}(i,j) = q^t(i,j)$ $\forall$ $i, j \text{ such that } \beta^t_i, \beta^t_j \neq G$ and $q^{t+1}(i,i_G) = q^{t+1}(i_G,i) = L(G,\beta^t_i)$, where $i_G$ denotes the location of $G$ in $B^{t+1}$ and      \begin{equation*}
     L(G,\beta) = \left\{
     \begin{array}{cc}
     \underset{\mathbf{x}_i \in G,\mathbf{x}_j \in \beta}{\min} \delta(\mathbf{x}_i,\mathbf{x}_j) &\mbox{for SL-FWPD},\\
     \underset{\mathbf{x}_i \in G,\mathbf{x}_j \in \beta}{\max} \delta(\mathbf{x}_i,\mathbf{x}_j) &\mbox{for CL-FWPD},\\
     \frac{1}{|G| \times |\beta|} \underset{\mathbf{x}_i \in G}{\sum} \underset{\mathbf{x}_j \in \beta}{\sum} \delta(\mathbf{x}_i,\mathbf{x}_j) &\mbox{for AL-FWPD}.\\
     \end{array}\right.
     \end{equation*}
     Set $t=t+1$.
 \item Repeat Steps 2-4 until $B^t$ contains a single element.
\end{enumerate}

FWPD being the underlying dissimilarity measure (instead of other metrics such as the Euclidean distance), the HAC-FWPD algorithm can be directly applied to obtain SL, CL, or AL based hierarchical clustering of datasets with missing feature values.

\subsection{Time Complexity of the HAC-FWPD Algorithm}\label{sec:TimeCompHierFwpd}

Irrespective of whether SL-FWPD, AL-SWPD or CL-FWPD is used as the proximity measure, the HAC-FWPD algorithm consists of the following three basic steps:
\begin{enumerate}
 \item \emph{Distance Calculation}: As each distance calculation involves at most $m$ features, the calculation of all pairwise observed distances among $n$ data instances is at most $\mathcal{O}(n^{2}m)$.
 \item \emph{Penalty Calculation}: The penalty calculation between a data point and a cluster centroid involves at most $m$ summations. Hence, penalty calculation over all possible pairings is at most $\mathcal{O}(n^{2}m)$.
 \item \emph{Cluster Merging}: The merging of two clusters takes place in each of the $n-1$ steps of the algorithm, and each merge at most has a time complexity of $\mathcal{O}(n^2)$.
\end{enumerate}
Therefore, assuming $m \ll n$, the total computational complexity of the HAC-FWPD algorithm is $\mathcal{O}(n^{3})$ which is the same as that of the standard HAC algorithm based on SL, CL or AL.

\section{Experimental Results}\label{sec:ExpRes}

In this section, we report the results of several experiments carried out to validate the merit of the proposed k-means-FWPD and HAC-FWPD clustering algorithms. In the following subsections, we describe the experimental setup used to validate the proposed techniques. The results of the experiments for the k-means-FWPD algorithm and the HAC-FWPD algorithm, are respectively presented thereafter.

\subsection{Experiment Setup}\label{sec:ExpPhil}

Adjusted Rand Index (ARI) \cite{hubert1985comparing} is a popular validity index used to judge the merit of the clustering algorithms. When the true class labels are known, ARI provides a measure of the similarity between the cluster partition obtained by a clustering technique and the true class labels. Therefore, a high value of ARI is thought to indicate better clusterings. But, the class labels may not always be in keeping with the natural cluster structure of the dataset. In such cases, good clusterings are likely to achieve lower values of these indexes compared to possibly erroneous partitions (which are more akin to the class labels). However, the purpose of our experiments is to find out how close the clusterings obtained by the proposed methods (and the contending techniques) are to the clusterings obtained by the standard algorithms (k-means algorithm and HAC algorithm); the proposed methods (and its contenders) being run on the datasets with missingness, while the standard methods are run on corresponding fully observed datasets. Hence, the clusterings obtained by the standard algorithms are used as the ground-truths using which the ARI values are calculated for the proposed methods (and their contenders). The performances of ZI, MI, kNNI (with $k \in \{3,5,10,20\}$) and SVDI (using the most significant 10\% of the eigenvectors) are used for comparison with the proposed methods. The variant of MI that we impute with for these experiments differs from the traditional technique in that we use the average of the averages for individual classes, instead of the overall average. This is done to minimize the effects of severe class imbalances that may exist in the datasets. We also conduct the Wilcoxon's signed rank test \cite{wilcoxon1945individual} to evaluate the statistical significance of the observed results.

The performance of k-means depends on the initial cluster assignment. Therefore, to ensure fairness, we use the same set of random initial cluster assignments for both the standard k-means algorithm on the fully observed dataset as well as the proposed k-means-FWPD method (and its contenders). The maximum number of iterations of the k-means variants is set as $MaxIter = 500$. Results are recorded in terms of average ARI values over 50 different runs on each dataset. The number of clusters is assumed to be same as the number of classes.

For HAC experiments, Results are reported as average ARI values obtained over 20 independent runs. AL is chosen over SL and CL as it is observed to generally achieve higher ARI values. The number of clusters is assumed to be same as the number of classes.

\subsubsection{Datasets}

We take 20 real-world datasets from the University of California at Irvine (UCI) repository \cite{dheeru2017uci} and the Jin Genomics Dataset (JGD) repository \cite{jin2017data}. Each feature of each dataset is normalized so as to have zero mean and unit standard deviation. The details of these 20 datasets are listed in Table \ref{dataDetails}.

\subsubsection{Simulating Missingness Mechanisms}\label{mechanism}

Experiments are conducted by removing features from each of the datasets according to the four missingness mechanisms, namely MCAR, MAR, MNAR-I and MNAR-II \cite{datta2016fwpd}. The detailed algorithm for simulating the four missingness mechanisms is as follows:

\begin{enumerate}
\item Specify the number of entries $MissCount$ to be removed from the dataset. Select the missingness mechanism as one out of MCAR, MAR, MNAR-I or MNAR-II.
\item If the mechanism is MAR or MNAR-II, select a random subset $\gamma_{miss} \subset S$ containing half of the features in $S$ (i.e. $|\gamma_{miss}| = \frac{m}{2}$ if $|S|$ is even or $\frac{m+1}{2}$ if $|S|$ is odd). If the mechanism is MNAR-I, set $\gamma_{miss} = S$. Identify $\gamma_{obs} = S \backslash \gamma_{miss}$. Otherwise, go to Step 5.
\item If the mechanism is MAR or MNAR-II, for each feature $l \in \gamma_{miss}$, randomly select a feature $l_c \in \gamma_{obs}$ on which the missingness of feature $l$ may depend.
\item For each feature $l \in \gamma_{miss}$ randomly choose a type of missingness $MissType_l$ as one out of CENTRAL, INTERMEDIATE or EXTREMAL.
\item Randomly select a non-missing entry $x_{i,l}$ from the data matrix. If the mechanism is MCAR, mark the entry as missing and decrement $MissCount = MissCount - 1$ and go to Step 11.
\item If the mechanism is MAR, set $\lambda = x_{i,l_c}$, $\mu = \mu_{l_c}$ and $\sigma = \sigma_{l_c}$, where $\mu_{l_c}$ and $\sigma_{l_c}$ are the mean and standard deviation of the $l_c$-th feature over the dataset. If the mechanism is MNAR-I, set $\lambda = x_{i,l}$, $\mu = \mu_{l}$ and $\sigma = \sigma_{l}$. If the mechanism is MNAR-II, randomly set either $\lambda = x_{i,l}$, $\mu = \mu_{l}$ and $\sigma = \sigma_{l}$ or $\lambda = x_{i,l_c}$, $\mu = \mu_{l_c}$ and $\sigma = \sigma_{l_c}$.
\item Calculate $z = \frac{1}{\sigma}\sqrt{(\lambda - \mu)^2}$.
\item If $MissType_l = \text{CENTRAL}$, set $\mu_z = 0$. If $MissType_l = \text{INTERMEDIATE}$, set $\mu_z = 1$. If $MissType_l = \text{EXTREMAL}$, set $\mu_z = 2$. Set $\sigma_z = 0.35$.
\item Calculate $pval = \frac{1}{\sqrt{2 \pi \sigma_z}} exp(- \frac{(z - \mu_z)^2}{2 \sigma_z^2})$.
\item Randomly generate a value $qval$ in the interval $[0,1]$. If $pval \geq qval$, then mark the entry $x_{i,l}$ as missing and decrement $MissCount = MissCount - 1$.
\item If $MissCount > 0$, then go to Step 5.
\end{enumerate}

In the above algorithm, the dependence of the missingness on feature values for MAR, MNAR-I and MNAR-II is achieved by removing entries based on the values of control features for their corresponding data points. The control feature may be the feature itself (for MNAR-I and MNAR-II) or may be another feature for the same data point (as in the case of MAR and MNAR-II). The dependence can be of three types, namely CENTRAL, INTERMEDIATE or EXTREMAL. CENTRAL dependence removes a feature when its corresponding control feature has a value close to the mean of the control feature over the dataset. EXTREMAL dependence removes a feature when the value of its control feature lies near the extremes. INTERMEDIATE dependence removes a feature when the value of the control lies between the mean and the extremes.

For our experiments, we set $MissCount = \frac{nm}{4}$ to remove 25\% of the features from each dataset. Thus, an average of $\frac{m}{4}$ of the features are missing from each data instance.

\begin{table}[h]
\begin{center}
\caption{Details of the 20 real-world datasets}\label{dataDetails}
\begin{tabular}{c c c c c}
\hline
Dataset & \#Instances & \#Features & \#Classes & Repository\\
\hline
Chronic Kidney & 800 & 24 & 2 & UCI\\
Colon & 62 & 2000 & 2 & JGD\\
GSAD$^{*}$ 1$^{\dagger}$ & 445 & 128 & 6 & UCI\\
Glass & 214 & 9 & 6 & UCI\\
Iris & 150 & 4 & 3 & UCI\\
Isolet 5$^{\dagger}$ & 1559 & 617 & 26 & UCI\\
Landsat & 6435 & 36 & 6 & UCI\\
Leaf & 340 & 15 & 36 & UCI\\
Libras & 360 & 90 & 15 & UCI\\
Lung & 181 & 12533 & 2 & JGD\\
Lung Cancer & 27 & 56 & 3 & UCI\\
Lymphoma & 62 & 4026 & 3 & JGD\\
Pendigits & 10992 & 16 & 10 & UCI\\
Prostate & 102 & 6033 & 2 & JGD\\
Seeds & 210 & 7 & 3 & UCI\\
Sensorless$^{\dagger}$ & 6000 & 48 & 11 & UCI\\
Sonar & 208 & 60 & 2 & UCI\\
Theorem Proving$^{\dagger}$ & 3059 & 51 & 6 & UCI\\
Vehicle & 94 & 18 & 4 & UCI\\
Vowel Context & 990 & 14 & 11 & UCI\\
\hline
\multicolumn{5}{l}{\footnotesize{$^{*}$Gas Sensor Array Drift}}\\
\multicolumn{5}{l}{\footnotesize{$^{\dagger}$Only a meaningful subset of the dataset is used.}}\\
\end{tabular}
\end{center}
\end{table}

\subsubsection{Selecting the parameter $\alpha$}\label{sec:alpha}

In order to conduct experiments using the FWPD measure, we need to select a value of the parameter $\alpha$. Proper selection of $\alpha$ may help to boost the performance of the proposed k-means-FWPD and HAC-FWPD measures. Therefore, in this section, we undertake a study on the effects of $\alpha$ on the performance of FWPD. Experiments are conducted using $\alpha \in \{0.1, 0.25, 0.5, 0.75, 0.9\}$ on the datasets listed in Table \ref{dataDetails} using the experimental setup detailed above. The summary of the results of this study is shown in Table \ref{tabAlphas} in terms of average ARI values. 

A choice of $\alpha = 0.25$ performs best overall as well as individually except for MAR missingness (where $\alpha=0.1$ proves to be a better choice). This seems to indicate some correlation between the extent of missingness and the optimal value of $\alpha$ (25\% of the features are missing in our experiments as mentioned in Section \ref{mechanism}). However, the correlation is rather weak for k-means-FWPD where all values of alphas seem to have competitive performance. On the other hand, the correlation is seen to be much stronger for HAC-FWPD. This indicates that the optimal $\alpha$ varies considerably with the pattern of missingness for k-means-FWPD but not as much for HAC-FWPD. Another interesting observation is that performance of HAC-FWPD deteriorates considerably for high values of $\alpha$ implying that the distance term is FWPD must be given greater importance for HAC methods. As $\alpha=0.25$ has the best performance overall, we report the detailed experimental results in the subsequent sections for this choice of $\alpha$.

\begin{table}[h]
\begin{center}
\caption{Summary of results for different choices of $\alpha$ in terms of average ARI values.}\label{tabAlphas}
\scriptsize{
\begin{tabular}{c | c | c c c c c}
\hline
Clustering & Type of & \multicolumn{5}{c}{$\alpha$}\\
Algorithm & Missingness & 0.1 & 0.25 & 0.5 & 0.75 & 0.9\\
\hline
k-means- & MCAR & 0.682 & 0.712 & 0.664 & 0.691 & 0.683 \\
-FWPD & MAR & 0.738 & 0.730 & 0.723 & 0.729 & 0.711 \\
 & MNAR-I & 0.649 & 0.676 & 0.675 & 0.613 & 0.666 \\
 & MNAR-II & 0.711 & 0.718 & 0.689 & 0.665 & 0.678 \\
\cline{2-7}
 & Overall & 0.695 & \textbf{0.709} & 0.688 & 0.675 & 0.685 \\
\hline
HAC-FWPD & MCAR & 0.665 & 0.709 & 0.389 & 0.073 & 0.017 \\
 & MAR & 0.740 & 0.724 & 0.441 & 0.210 & 0.094 \\
 & MNAR-I & 0.720 & 0.721 & 0.458 & 0.158 & 0.036 \\
 & MNAR-II & 0.708 & 0.716 & 0.443 & 0.140 & 0.025 \\
\cline{2-7}

 & Overall & 0.709 & \textbf{0.718} & 0.433 & 0.145 & 0.043 \\
\hline
\multicolumn{7}{l}{Best values shown in \textbf{boldface}.}\\
\end{tabular}
}
\end{center}
\end{table}

\subsection{Experiments with the k-means-FWPD Algorithm}\label{sec:kmeansExps}

We compare the proposed k-means-FWPD algorithm to the standard k-means algorithm run on the datasets obtained after performing ZI, MI, SVDI and kNNI. All runs of k-means-FWPD were found to converge within the stipulated budget of $MaxIter = 500$. The results of the experiments are listed in terms of the means and standard deviations of the obtained ARI values, in Tables \ref{mcarKmResults}-\ref{mnar2KmResults}. Only the best results for kNNI are reported, along with the best $k$ values. The statistically significance of the listed results are summarized at the bottom of the table in terms of average ranks as well as signed rank test hypotheses and p-values ($H_0$ signifying that the ARI values achieved by the proposed method and the contending method originate from identical distributions having the same medians; $H_1$ implies that the ARI values achieved by the proposed method and the contender originate from different distributions).

\begin{table*}[!t]
\begin{center}
\caption{Means and standard deviations of ARI values for the k-means-FWPD algorithm against MCAR.}\label{mcarKmResults}
\scriptsize{
\begin{tabular}{c c c c c c c}
\hline
Dataset & $\begin{array}{ll}
\text{k-means-}\\
\text{-FWPD}\\
\end{array}$ & ZI & MI & SVDI & kNNI & $\begin{array}{cc}
\text{Best } k\\
\text{(for kNNI)}\\
\end{array}$\\
\hline
Chronic Kidney & 0.807$\pm$0.002 & 0.813$\pm$0.005 & 0.229$\pm$0.013 & 0.763$\pm$0.005 & \textbf{0.815}$\pm$0.003 & 5 \\
Colon 		   & \textbf{0.681}$\pm$0.341 & 0.656$\pm$0.323 & 0.659$\pm$0.322 & 0.662$\pm$0.314 & 0.656$\pm$0.323 & 3 \\
GSAD 1 		   & \textbf{0.798}$\pm$0.236 & 0.625$\pm$0.199 & 0.722$\pm$0.172 & 0.552$\pm$0.203 & 0.711$\pm$0.195 & 3 \\
Glass          & 0.488$\pm$0.097 & 0.466$\pm$0.119 & 0.131$\pm$0.066 & 0.417$\pm$0.114 & \textbf{0.505}$\pm$0.134 & 5 \\
Iris           & \textbf{0.799}$\pm$0.119 & 0.672$\pm$0.113 & 0.116$\pm$0.083 & 0.732$\pm$0.159 & 0.758$\pm$0.157 & 3 \\
Isolet 5 	   & \textbf{0.679}$\pm$0.117 & 0.623$\pm$0.093 & 0.625$\pm$0.097 & 0.626$\pm$0.105 & 0.614$\pm$0.072 & 3 \\
Landsat 	   & \textbf{0.937}$\pm$0.001 & 0.807$\pm$0.001 & 0.798$\pm$0.001 & 0.838$\pm$0.104 & 0.937$\pm$0.010 & 5 \\
Leaf 		   & 0.455$\pm$0.014 & 0.328$\pm$0.010 & 0.339$\pm$0.019 & 0.354$\pm$0.037 & \textbf{0.465}$\pm$0.029 & 5 \\
Libras         & \textbf{0.656}$\pm$0.070 & 0.642$\pm$0.069 & 0.103$\pm$0.019 & 0.619$\pm$0.067 & 0.625$\pm$0.077 & 20\\
Lung 		   & \textbf{0.731}$\pm$0.341 & 0.718$\pm$0.261 & 0.659$\pm$0.341 & 0.694$\pm$0.318 & 0.718$\pm$0.261 & 3 \\
Lung Cancer    & \textbf{0.542}$\pm$0.249 & 0.541$\pm$0.202 & 0.537$\pm$0.214 & 0.529$\pm$0.192 & 0.525$\pm$0.189 & 5 \\
Lymphoma 	   & \textbf{0.755}$\pm$0.167 & 0.743$\pm$0.175 & 0.700$\pm$0.165 & 0.733$\pm$0.175 & 0.743$\pm$0.175 & 3 \\
Pendigits      & 0.729$\pm$0.089 & 0.659$\pm$0.082 & 0.083$\pm$0.013 & 0.604$\pm$0.063 & \textbf{0.832}$\pm$0.105 & 3 \\
Prostate 	   & \textbf{0.961}$\pm$0.025 & 0.944$\pm$0.043 & 0.944$\pm$0.043 & 0.946$\pm$0.041 & 0.944$\pm$0.043 & 3 \\
Seeds          & \textbf{0.866}$\pm$0.030 & 0.735$\pm$0.021 & 0.242$\pm$0.039 & 0.745$\pm$0.041 & 0.865$\pm$0.025 & 5 \\
Sensorless & \textbf{0.765}$\pm$0.031 & 0.684$\pm$0.028 & 0.687$\pm$0.021 & 0.719$\pm$0.060 & 0.726$\pm$0.051 & 20 \\
Sonar 	   & \textbf{0.697}$\pm$0.195 & 0.681$\pm$0.187 & 0.672$\pm$0.188 & 0.434$\pm$0.234 & 0.656$\pm$0.162 & 5 \\
Theorem Proving & \textbf{0.714}$\pm$0.197 & 0.671$\pm$0.229 & 0.661$\pm$0.188 & 0.565$\pm$0.139 & 0.672$\pm$0.197 & 20 \\
Vehicle        & 0.715$\pm$0.143 & 0.674$\pm$0.139 & 0.114$\pm$0.060 & 0.646$\pm$0.105 & \textbf{0.723}$\pm$0.134 & 10\\
Vowel Context  & 0.458$\pm$0.031 & 0.366$\pm$0.028 & 0.360$\pm$0.029 & 0.352$\pm$0.022 & \textbf{0.461}$\pm$0.060 & 3 \\
\hline
Average Ranks  & \textbf{1.38} & 3.33 & 4.20 & 3.65 & 2.45 &   \\
\hline
\multicolumn{2}{c}{Signed Rank Hypotheses (p-values)} & $H_1 (0.00)$ & $H_1 (0.00)$ & $H_1 (0.00)$ & $H_1 (0.03)$ & \\
\hline
\multicolumn{7}{l}{$H_1$ significantly different from k-means-FWPD}\\
\multicolumn{7}{l}{$H_0$ statistically similar to k-means-FWPD}\\
\multicolumn{7}{l}{Best values shown in \textbf{boldface}.}\\
\end{tabular}
}
\end{center}
\end{table*}

\begin{table*}[!t]
\begin{center}
\caption{Means and standard deviations of ARI values for the k-means-FWPD algorithm against MAR.}\label{marKmResults}
\scriptsize{
\begin{tabular}{c c c c c c c}
\hline
Dataset & $\begin{array}{ll}
\text{k-means-}\\
\text{-FWPD}\\
\end{array}$ & ZI & MI & SVDI & kNNI & $\begin{array}{cc}
\text{Best } k\\
\text{(for kNNI)}\\
\end{array}$\\
\hline
Chronic Kidney & 0.793$\pm$0.006 & 0.792$\pm$0.004 & 0.803$\pm$0.003 & 0.787$\pm$0.008 & \textbf{0.838}$\pm$0.011 & 20\\
Colon 		   & 0.812$\pm$0.177 & 0.826$\pm$0.149 & \textbf{0.827}$\pm$0.157 & 0.796$\pm$0.202 & 0.826$\pm$0.149 & 3 \\
GSAD 1 		   & \textbf{0.801}$\pm$0.163 & 0.737$\pm$0.187 & 0.673$\pm$0.179 & 0.719$\pm$0.165 & 0.760$\pm$0.144 & 3 \\
Glass          & \textbf{0.617}$\pm$0.125 & 0.455$\pm$0.168 & 0.565$\pm$0.128 & 0.411$\pm$0.171 & 0.479$\pm$0.152 & 5 \\
Iris           & 0.776$\pm$0.185 & 0.776$\pm$0.185 & \textbf{0.851}$\pm$0.144 & 0.776$\pm$0.163 & 0.764$\pm$0.176 & 5 \\
Isolet 5 	   & \textbf{0.729}$\pm$0.072 & 0.713$\pm$0.051 & 0.691$\pm$0.046 & 0.704$\pm$0.027 & 0.713$\pm$0.051 & 3 \\
Landsat 	   & \textbf{0.940}$\pm$0.002 & 0.828$\pm$0.127 & 0.850$\pm$0.123 & 0.773$\pm$0.150 & 0.899$\pm$0.058 & 3 \\
Leaf 		   & 0.510$\pm$0.022 & 0.392$\pm$0.042 & 0.440$\pm$0.046 & 0.501$\pm$0.021 & \textbf{0.532}$\pm$0.046 & 3 \\
Libras         & 0.731$\pm$0.076 & 0.700$\pm$0.077 & \textbf{0.778}$\pm$0.065 & 0.697$\pm$0.082 & 0.675$\pm$0.071 & 3 \\
Lung 		   & \textbf{0.754}$\pm$0.131& 0.711$\pm$0.230 & 0.717$\pm$0.232 & 0.625$\pm$0.312 & 0.711$\pm$0.230 & 3 \\
Lung Cancer    & \textbf{0.606}$\pm$0.223 & 0.526$\pm$0.224 & 0.476$\pm$0.219 & 0.509$\pm$0.202 & 0.493$\pm$0.231 & 3 \\
Lymphoma 	   & 0.790$\pm$0.164 & 0.883$\pm$0.109 & \textbf{0.885}$\pm$0.102 & 0.875$\pm$0.120 & 0.883$\pm$0.109 & 3 \\
Pendigits      & 0.717$\pm$0.068 & 0.494$\pm$0.072 & 0.852$\pm$0.062 & 0.705$\pm$0.067 & \textbf{0.903}$\pm$0.065 & 5 \\
Prostate 	   & \textbf{0.990}$\pm$0.021 & 0.984$\pm$0.036 & 0.986$\pm$0.036 & 0.918$\pm$0.017 & 0.984$\pm$0.036 & 3 \\
Seeds          & 0.785$\pm$0.026 & 0.755$\pm$0.025 & 0.774$\pm$0.027 & 0.752$\pm$0.025 & \textbf{0.834}$\pm$0.033 & 3 \\
Sensorless     & \textbf{0.759}$\pm$0.038 & 0.663$\pm$0.089 & 0.629$\pm$0.145 & 0.662$\pm$0.132 & 0.685$\pm$0.087 & 3 \\
Sonar 		   & \textbf{0.620}$\pm$0.289 & 0.599$\pm$0.326 & 0.598$\pm$0.325 & 0.524$\pm$0.315 & 0.574$\pm$0.359 & 10 \\
Theorem Proving & \textbf{0.672}$\pm$0.179 & 0.636$\pm$0.165 & 0.617$\pm$0.182 & 0.618$\pm$0.189 & 0.649$\pm$0.145 & 10 \\
Vehicle        & \textbf{0.699}$\pm$0.142 & 0.545$\pm$0.155 & 0.644$\pm$0.147 & 0.566$\pm$0.154 & 0.537$\pm$0.149 & 3 \\
Vowel Context  & 0.497$\pm$0.044 & 0.461$\pm$0.072 & 0.436$\pm$0.064 & 0.408$\pm$0.054 & \textbf{0.577}$\pm$0.043 & 3 \\
\hline
Average Ranks  & \textbf{1.85} & 3.33 & 2.90 & 4.25 & 2.67 &   \\
\hline
\multicolumn{2}{c}{Signed Rank Hypotheses (p-values)} & $H_1 (0.00)$ & $H_0 (0.13)$ & $H_1 (0.00)$ & $H_0 (0.37)$ & \\
\hline
\multicolumn{7}{l}{$H_1$ significantly different from k-means-FWPD}\\
\multicolumn{7}{l}{$H_0$ statistically similar to k-means-FWPD}\\
\multicolumn{7}{l}{Best values shown in \textbf{boldface}.}\\
\end{tabular}
}
\end{center}
\end{table*}

\begin{table*}[!t]
\begin{center}
\caption{Means and standard deviations of ARI values for the k-means-FWPD algorithm against MNAR-I.}\label{mnar1KmResults}
\scriptsize{
\begin{tabular}{c c c c c c c}
\hline
Dataset & $\begin{array}{ll}
\text{k-means-}\\
\text{-FWPD}\\
\end{array}$ & ZI & MI & SVDI & kNNI & $\begin{array}{cc}
\text{Best } k\\
\text{(for kNNI)}\\
\end{array}$\\
\hline
Chronic Kidney & \textbf{0.729}$\pm$0.011 & 0.714$\pm$0.010 & 0.399$\pm$0.051 & 0.599$\pm$0.005 & 0.616$\pm$0.022 & 3 \\
Colon 		   & 0.789$\pm$0.214 & 0.781$\pm$0.202 & 0.770$\pm$0.205 & \textbf{0.801}$\pm$0.147 & 0.781$\pm$0.202 & 3 \\
GSAD 1 		   & 0.791$\pm$0.112 & \textbf{0.799}$\pm$0.097 & 0.790$\pm$0.110 & 0.689$\pm$0.175 & \textbf{0.799}$\pm$0.097 & 3 \\
Glass          & \textbf{0.439}$\pm$0.101 & 0.391$\pm$0.117 & 0.152$\pm$0.048 & 0.388$\pm$0.097 & 0.438$\pm$0.105 & 10\\
Iris           & 0.662$\pm$0.073 & 0.709$\pm$0.144 & 0.137$\pm$0.077 & \textbf{0.739}$\pm$0.132 & 0.658$\pm$0.168 & 5 \\
Isolet 5 	   & \textbf{0.708}$\pm$0.098 & 0.680$\pm$0.103 & 0.680$\pm$0.082 & 0.663$\pm$0.067 & 0.680$\pm$0.103 & 3 \\
Landsat 	   & \textbf{0.869}$\pm$0.048 & 0.701$\pm$0.149 & 0.712$\pm$0.159 & 0.858$\pm$0.058 & 0.813$\pm$0.001 & 10 \\
Leaf 		   & 0.493$\pm$0.052 & 0.416$\pm$0.029 & 0.403$\pm$0.020 & 0.439$\pm$0.023 & \textbf{0.522}$\pm$0.040 & 3 \\
Libras         & \textbf{0.717}$\pm$0.083 & 0.667$\pm$0.076 & 0.378$\pm$0.058 & 0.638$\pm$0.070 & 0.656$\pm$0.067 & 3 \\
Lung 		   & \textbf{0.636}$\pm$0.201 & 0.592$\pm$0.199 & 0.606$\pm$0.210 & 0.578$\pm$0.192 & 0.592$\pm$0.199 & 3 \\
Lung Cancer    & \textbf{0.529}$\pm$0.235 & 0.497$\pm$0.184 & 0.459$\pm$0.129 & 0.457$\pm$0.217 & 0.497$\pm$0.184 & 3 \\
Lymphoma 	   & 0.796$\pm$0.118 & \textbf{0.798}$\pm$0.133 & 0.764$\pm$0.130 & 0.764$\pm$0.129 & \textbf{0.798}$\pm$0.133 & 3 \\
Pendigits      & 0.666$\pm$0.079 & 0.635$\pm$0.067 & 0.135$\pm$0.025 & 0.619$\pm$0.054 & \textbf{0.756}$\pm$0.093 & 5 \\
Prostate 	   & \textbf{0.975}$\pm$0.032 & 0.958$\pm$0.075 & 0.962$\pm$0.075 & 0.910$\pm$0.061 & 0.958$\pm$0.075 & 3 \\
Seeds          & 0.776$\pm$0.019 & 0.705$\pm$0.044 & 0.298$\pm$0.065 & 0.725$\pm$0.042 & \textbf{0.819}$\pm$0.052 & 20\\
Sensorless 	   & \textbf{0.693}$\pm$0.041 & 0.638$\pm$0.036 & 0.636$\pm$0.039 & 0.610$\pm$0.069 & 0.593$\pm$0.051 & 10 \\
Sonar 		   & \textbf{0.600}$\pm$0.297 & 0.537$\pm$0.287 & 0.546$\pm$0.292 & 0.326$\pm$0.282 & 0.537$\pm$0.287 & 3 \\
Theorem Proving & \textbf{0.540}$\pm$0.211 & 0.399$\pm$0.196 & 0.388$\pm$0.178 & 0.513$\pm$0.224 & 0.465$\pm$0.195 & 3 \\
Vehicle        & \textbf{0.639}$\pm$0.141 & 0.551$\pm$0.123 & 0.298$\pm$0.078 & 0.613$\pm$0.103 & 0.536$\pm$0.107 & 3 \\
Vowel Context  & 0.473$\pm$0.044 & 0.412$\pm$0.049 & 0.435$\pm$0.056 & 0.383$\pm$0.043 & \textbf{0.512}$\pm$0.036 & 3 \\
\hline
Average Ranks  & \textbf{1.55} & 3.02 & 4.08 & 3.67 & 2.67 &   \\
\hline
\multicolumn{2}{c}{Signed Rank Hypotheses (p-values)} & $H_1 (0.00)$ & $H_1 (0.00)$ & $H_1 (0.00)$ & $H_1 (0.05)$ & \\
\hline
\multicolumn{7}{l}{$H_1$ significantly different from k-means-FWPD}\\
\multicolumn{7}{l}{$H_0$ statistically similar to k-means-FWPD}\\
\multicolumn{7}{l}{Best values shown in \textbf{boldface}.}\\
\end{tabular}
}
\end{center}
\end{table*}

\begin{table*}[!t]
\begin{center}
\caption{Means and standard deviations of ARI values for the k-means-FWPD algorithm against MNAR-II.}\label{mnar2KmResults}
\scriptsize{
\begin{tabular}{c c c c c c c}
\hline
Dataset & $\begin{array}{ll}
\text{k-means-}\\
\text{-FWPD}\\
\end{array}$ & ZI & MI & SVDI & kNNI & $\begin{array}{cc}
\text{Best } k\\
\text{(for kNNI)}\\
\end{array}$\\
\hline
Chronic Kidney & 0.751$\pm$0.014 & 0.659$\pm$0.032 & 0.744$\pm$0.015 & 0.636$\pm$0.037 & \textbf{0.770}$\pm$0.017 & 3 \\
Colon 		   & \textbf{0.804}$\pm$0.210 & 0.797$\pm$0.226 & 0.798$\pm$0.220 & 0.781$\pm$0.208 & 0.797$\pm$0.226 & 3 \\
GSAD 1 		   & \textbf{0.731}$\pm$0.198 & 0.665$\pm$0.242 & 0.712$\pm$0.212 & 0.689$\pm$0.215 & 0.665$\pm$0.242 & 3 \\
Glass          & \textbf{0.530}$\pm$0.089 & 0.423$\pm$0.101 & 0.413$\pm$0.099 & 0.395$\pm$0.109 & 0.451$\pm$0.101 & 5 \\
Iris           & \textbf{0.773}$\pm$0.165 & 0.718$\pm$0.172 & 0.635$\pm$0.189 & 0.702$\pm$0.174 & 0.756$\pm$0.175 & 10\\
Isolet 5 	   & \textbf{0.789}$\pm$0.061 & 0.765$\pm$0.076 & 0.747$\pm$0.056 & 0.728$\pm$0.063 & 0.765$\pm$0.076 & 3 \\
Landsat 	   & \textbf{0.892}$\pm$0.083 & 0.871$\pm$0.084 & 0.868$\pm$0.083 & 0.794$\pm$0.130 & 0.836$\pm$0.083 & 3 \\
Leaf 		   & \textbf{0.476}$\pm$0.021 & 0.385$\pm$0.036 & 0.381$\pm$0.031 & 0.389$\pm$0.033 & 0.454$\pm$0.028 & 3 \\
Libras         & \textbf{0.698}$\pm$0.079 & 0.675$\pm$0.078 & 0.681$\pm$0.077 & 0.648$\pm$0.080 & 0.669$\pm$0.081 & 3 \\
Lung 		   & 0.686$\pm$0.220 & 0.649$\pm$0.226 & \textbf{0.707}$\pm$0.089 & 0.674$\pm$0.216 & 0.649$\pm$0.226 & 3 \\
Lung Cancer    & \textbf{0.641}$\pm$0.200 & 0.640$\pm$0.283 & 0.567$\pm$0.243 & 0.568$\pm$0.190 & 0.640$\pm$0.283 & 3 \\
Lymphoma 	   & \textbf{0.856}$\pm$0.115 & 0.824$\pm$0.136 & 0.842$\pm$0.123 & 0.818$\pm$0.153 & 0.824$\pm$0.136 & 3 \\
Pendigits      & 0.608$\pm$0.082 & 0.589$\pm$0.095 & 0.561$\pm$0.097 & 0.557$\pm$0.096 & \textbf{0.825}$\pm$0.083 & 5 \\
Prostate 	   & \textbf{0.984}$\pm$0.032 & \textbf{0.984}$\pm$0.032 & \textbf{0.984}$\pm$0.032 & 0.969$\pm$0.053 & \textbf{0.984}$\pm$0.032 & 3 \\
Seeds          & \textbf{0.884}$\pm$0.028 & 0.738$\pm$0.039 & 0.772$\pm$0.038 & 0.771$\pm$0.038 & 0.831$\pm$0.029 & 10\\
Sensorless 	   & \textbf{0.747}$\pm$0.041 & 0.667$\pm$0.130 & 0.727$\pm$0.030 & 0.704$\pm$0.056 & 0.698$\pm$0.068 & 3 \\
Sonar 		   & \textbf{0.704}$\pm$0.227 & 0.662$\pm$0.235 & 0.658$\pm$0.221 & 0.314$\pm$0.152 & 0.662$\pm$0.235 & 3 \\
Theorem Proving & \textbf{0.640}$\pm$0.141 & 0.600$\pm$0.105 & 0.605$\pm$0.077 & 0.610$\pm$0.175 & 0.593$\pm$0.106 & 3 \\
Vehicle        & 0.677$\pm$0.145 & \textbf{0.734}$\pm$0.132 & 0.571$\pm$0.167 & 0.635$\pm$0.153 & 0.712$\pm$0.141 & 3 \\
Vowel Context  & 0.478$\pm$0.048 & 0.404$\pm$0.041 & 0.396$\pm$0.064 & 0.345$\pm$0.032 & \textbf{0.511}$\pm$0.056 & 3
 \\
\hline
Average Ranks  & \textbf{1.38} & 3.30 & 3.27 & 4.25 & 2.80 &   \\
\hline
\multicolumn{2}{c}{Signed Rank Hypotheses (p-values)} & $H_1 (0.00)$ & $H_1 (0.00)$ & $H_1 (0.00)$ & $H_1 (0.03)$ & \\
\hline
\multicolumn{7}{l}{$H_1$ significantly different from k-means-FWPD}\\
\multicolumn{7}{l}{$H_0$ statistically similar to k-means-FWPD}\\
\multicolumn{7}{l}{Best values shown in \textbf{boldface}.}\\
\end{tabular}
}
\end{center}
\end{table*}

We know from Theorem \ref{thmFinFin} that the maximum number of feasibility adjustments that can occur during a single run of k-means-FWPD is $n(k-1)$. This begs the question of whether one should choose $MaxIter \geq n(k-1)$. However, k-means-FWPD was observed to converge within the stipulated $MaxIter = 500$ iterations even for datasets like Isolet 5, Pendigits, Sensorless, etc. which have relatively large values of $n(k-1)$. This indicates that the number of feasibility adjustments that occur during a run is much lower in practice. Therefore, we conclude that it is not required to set $MaxIter \geq n(k-1)$ for practical problems.

It is seen from Tables \ref{mcarKmResults}-\ref{mnar2KmResults} that the k-means-FWPD algorithm performs best, indicated by the consistently minimum average rankings on all types of missingness. The proposed method performs best on the majority of datasets for all kinds of missingness. kNNI is overall seen to be the second best performer (being statistically comparable to k-means-FWPD in case of MAR). It is also interesting to observe that the performance of MI is improved in case of MAR and MNAR-II, indicating that MI tends to be useful for partitional clustering when the missingness depends on the observed features. Moreover, SVDI is generally observed to perform poorly irrespective of the type of missingness, implying that the linear model assumed by SVDI is unable to conserve the convexity of the clusters (which is essential for good performance in case of partitional clustering).

\subsection{Experiments with the HAC-FWPD Algorithm}\label{sec:hacExps}

The experimental setup described in Section \ref{sec:ExpPhil} is also used to compare the HAC-FWPD algorithm (with AL-FWPD as the proximity measure) to the standard HAC algorithm (with AL as the proximity measure) in conjunction with ZI, MI, SVDI and kNNI. Results are reported as means and standard deviations of obtained ARI values over the 20 independent runs. AL is preferred here over SL and CL as it is observed to generally achieve higher ARI values. The results of the experiments are listed in Tables \ref{mcarHacResults}-\ref{mnar2HacResults}. The statistically significance of the listed results are also summarized at the bottom of the respective tables in terms of average ranks as well as signed rank test hypotheses and p-values ($H_0$ signifying that the ARI values achieved by the proposed method and the contending method originate from identical distributions having the same medians; $H_1$ implies that the ARI values achieved by the proposed method and the contender originate from different distributions).

\begin{table*}[!t]
\begin{center}
\caption{Means and standard deviations of ARI values for the HAC-FWPD algorithm against MCAR.}\label{mcarHacResults}
\scriptsize{
\begin{tabular}{c c c c c c c}
\hline
Dataset & $\begin{array}{ll}
\text{HAC-}\\
\text{-FWPD}\\
\end{array}$ & ZI & MI & SVDI & kNNI & $\begin{array}{cc}
\text{Best } k\\
\text{(for kNNI)}\\
\end{array}$\\
\hline
Chronic Kidney & \textbf{1.000}$\pm$0.000 & 0.967$\pm$0.031 & 0.933$\pm$0.033 & 0.933$\pm$0.033 & 0.000$\pm$0.000 & 3 \\
Colon 		   & \textbf{0.690}$\pm$0.240 & 0.469$\pm$0.145 & 0.286$\pm$0.174 & 0.380$\pm$0.304 & 0.000$\pm$0.000 & 3 \\
GSAD 1 		   & \textbf{0.454}$\pm$0.309 & 0.367$\pm$0.088 & 0.271$\pm$0.112 & 0.311$\pm$0.087 & 0.022$\pm$0.018 & 3 \\
Glass          & \textbf{0.737}$\pm$0.081 & 0.671$\pm$0.089 & 0.680$\pm$0.089 & 0.638$\pm$0.090 & 0.033$\pm$0.032 & 3 \\
Iris           & 0.885$\pm$0.072 & \textbf{0.922}$\pm$0.047 & 0.831$\pm$0.053 & 0.917$\pm$0.049 & 0.559$\pm$0.129 & 20\\
Isolet 5 	   & \textbf{0.855}$\pm$0.111 & 0.044$\pm$0.003 & 0.046$\pm$0.003 & 0.081$\pm$0.037 & 0.064$\pm$0.003 & 3 \\
Landsat 	   & \textbf{0.712}$\pm$0.098 & 0.228$\pm$0.034 & 0.254$\pm$0.012 & 0.217$\pm$0.033 & 0.300$\pm$0.018 & 10 \\
Leaf 		   & \textbf{0.497}$\pm$0.046 & 0.200$\pm$0.016 & 0.221$\pm$0.017 & 0.290$\pm$0.077 & 0.140$\pm$0.011 & 3 \\
Libras         & \textbf{0.845}$\pm$0.054 & 0.276$\pm$0.031 & 0.298$\pm$0.033 & 0.381$\pm$0.030 & 0.156$\pm$0.050 & 10\\
Lung 		   & \textbf{1.000}$\pm$0.000 & \textbf{1.000}$\pm$0.000 & \textbf{1.000}$\pm$0.000 & \textbf{1.000}$\pm$0.000 & 0.001$\pm$0.002 & 3 \\
Lung Cancer    & \textbf{0.458}$\pm$0.193 & 0.408$\pm$0.223 & 0.356$\pm$0.229 & 0.436$\pm$0.335 & 0.034$\pm$0.035 & 3 \\
Lymphoma 	   & \textbf{0.885}$\pm$0.058 & 0.718$\pm$0.373 & 0.335$\pm$0.498 & 0.713$\pm$0.372 & 0.547$\pm$0.297 & 3 \\
Pendigits      & \textbf{0.712}$\pm$0.082 & 0.242$\pm$0.194 & 0.228$\pm$0.224 & 0.252$\pm$0.260 & 0.365$\pm$0.147 & 3 \\
Prostate 	   & \textbf{1.000}$\pm$0.000 & \textbf{1.000}$\pm$0.000 & \textbf{1.000}$\pm$0.000 & \textbf{1.000}$\pm$0.000 & 0.001$\pm$0.001 & 3 \\
Seeds          & 0.534$\pm$0.173 & 0.332$\pm$0.046 & 0.317$\pm$0.055 & 0.436$\pm$0.127 & \textbf{0.563}$\pm$0.110 & 10\\
Sensorless 	   & \textbf{0.416}$\pm$0.303 & 0.196$\pm$0.024 & 0.203$\pm$0.017 & 0.249$\pm$0.102 & 0.005$\pm$0.008 & 3 \\
Sonar 		   & \textbf{0.440}$\pm$0.419 & 0.329$\pm$0.473 & 0.128$\pm$0.298 & 0.261$\pm$0.365 & 0.001$\pm$0.000 & 3 \\
Theorem Proving & \textbf{0.802}$\pm$0.085 & 0.691$\pm$0.088 & 0.691$\pm$0.088 & 0.654$\pm$0.082 & 0.002$\pm$0.001 & 5 \\
Vehicle        & \textbf{0.807}$\pm$0.108 & 0.315$\pm$0.295 & 0.315$\pm$0.295 & 0.645$\pm$0.232 & 0.084$\pm$0.008 & 5 \\
Vowel Context  & \textbf{0.453}$\pm$0.081 & 0.248$\pm$0.042 & 0.211$\pm$0.066 & 0.194$\pm$0.029 & 0.101$\pm$0.019 & 3 \\
\hline
Average Ranks  & \textbf{1.30} & 2.95 & 3.52 & 2.88 & 4.35 &   \\
\hline
\multicolumn{2}{c}{Signed Rank Hypotheses (p-values)} & $H_1 (0.00)$ & $H_1 (0.00)$ & $H_1 (0.00)$ & $H_1 (0.00)$ & \\
\hline
\multicolumn{7}{l}{$H_1$ significantly different from HAC-FWPD}\\
\multicolumn{7}{l}{$H_0$ statistically similar to HAC-FWPD}\\
\multicolumn{7}{l}{Best values shown in \textbf{boldface}.}\\
\end{tabular}
}
\end{center}
\end{table*}

\begin{table*}[!t]
\begin{center}
\caption{Means and standard deviations of ARI values for the HAC-FWPD algorithm against MAR.}\label{marHacResults}
\scriptsize{
\begin{tabular}{c c c c c c c}
\hline
Dataset & $\begin{array}{ll}
\text{HAC-}\\
\text{-FWPD}\\
\end{array}$ & ZI & MI & SVDI & kNNI & $\begin{array}{cc}
\text{Best } k\\
\text{(for kNNI)}\\
\end{array}$\\
\hline
Chronic Kidney & \textbf{0.799}$\pm$0.394 & 0.398$\pm$0.494 & 0.398$\pm$0.494 & 0.398$\pm$0.494 & 0.003$\pm$0.001 & 3 \\
Colon 		   & \textbf{1.000}$\pm$0.000 & 0.463$\pm$0.516 & 0.463$\pm$0.516 & 0.601$\pm$0.423 & 0.016$\pm$0.002 & 3 \\
GSAD 1 		   & \textbf{0.619}$\pm$0.230 & 0.359$\pm$0.115 & 0.419$\pm$0.227 & 0.346$\pm$0.150 & 0.007$\pm$0.005 & 3 \\
Glass          & \textbf{0.650}$\pm$0.188 & 0.617$\pm$0.124 & 0.603$\pm$0.129 & 0.590$\pm$0.184 & 0.057$\pm$0.082 & 20 \\
Iris           & \textbf{0.949}$\pm$0.051 & 0.893$\pm$0.083 & 0.893$\pm$0.083 & 0.854$\pm$0.146 & 0.587$\pm$0.068 & 10 \\
Isolet 5 	   & \textbf{0.725}$\pm$0.186 & 0.491$\pm$0.011 & 0.491$\pm$0.011 & 0.464$\pm$0.076 & 0.076$\pm$0.005 & 3 \\
Landsat 	   & \textbf{0.734}$\pm$0.044 & 0.638$\pm$0.323 & 0.601$\pm$0.301 & 0.721$\pm$0.142 & 0.162$\pm$0.113 & 3 \\
Leaf 		   & 0.463$\pm$0.107 & 0.420$\pm$0.099 & 0.415$\pm$0.058 & \textbf{0.471}$\pm$0.053 & 0.154$\pm$0.013 & 3 \\
Libras         & \textbf{0.864}$\pm$0.098 & 0.855$\pm$0.048 & 0.815$\pm$0.081 & 0.813$\pm$0.064 & 0.469$\pm$0.012 & 3\\
Lung 		   & \textbf{1.000}$\pm$0.000 & \textbf{1.000}$\pm$0.000 & \textbf{1.000}$\pm$0.000 & \textbf{1.000}$\pm$0.000 & 0.008$\pm$0.024 & 3 \\
Lung Cancer    & \textbf{0.709}$\pm$0.270 & 0.522$\pm$0.438 & 0.538$\pm$0.422 & 0.443$\pm$0.358 & 0.036$\pm$0.019 & 3 \\
Lymphoma 	   & \textbf{1.000}$\pm$0.000 & 0.903$\pm$0.089 & 0.772$\pm$0.132 & 0.890$\pm$0.104 & 0.788$\pm$0.000 & 3 \\
Pendigits      & \textbf{0.493}$\pm$0.138 & 0.351$\pm$0.124 & 0.292$\pm$0.076 & 0.483$\pm$0.103 & 0.407$\pm$0.035 & 3 \\
Prostate 	   & \textbf{1.000}$\pm$0.000 & \textbf{1.000}$\pm$0.000 & \textbf{1.000}$\pm$0.000 & \textbf{1.000}$\pm$0.000 & 0.001$\pm$0.000 & 3 \\
Seeds          & \textbf{0.564}$\pm$0.131 & 0.487$\pm$0.016 & 0.444$\pm$0.103 & 0.535$\pm$0.224 & 0.556$\pm$0.135 & 3 \\
Sensorless 	   & \textbf{0.439}$\pm$0.288 & 0.276$\pm$0.163 & 0.174$\pm$0.053 & 0.256$\pm$0.136 & 0.000$\pm$0.000 & 10 \\
Sonar 		   & \textbf{0.396}$\pm$0.551 & 0.005$\pm$0.000 & 0.005$\pm$0.000 & 0.094$\pm$0.222 & 0.001$\pm$0.000 & 3 \\
Theorem Proving & \textbf{0.725}$\pm$0.102 & 0.677$\pm$0.031 & 0.685$\pm$0.107 & 0.641$\pm$0.121 & 0.001$\pm$0.006 & 3 \\
Vehicle        & \textbf{0.827}$\pm$0.123 & 0.431$\pm$0.279 & 0.431$\pm$0.278 & 0.825$\pm$0.105 & 0.075$\pm$0.044 & 3 \\
Vowel Context  & \textbf{0.517}$\pm$0.127 & 0.451$\pm$0.095 & 0.445$\pm$0.126 & 0.401$\pm$0.207 & 0.104$\pm$0.024 & 3 \\
\hline
Average Ranks  & \textbf{1.20} & 2.83 & 3.27 & 3.00 & 4.70 &   \\
\hline
\multicolumn{2}{c}{Signed Rank Hypotheses (p-values)} & $H_1 (0.00)$ & $H_1 (0.00)$ & $H_1 (0.00)$ & $H_1 (0.00)$ & \\
\hline
\multicolumn{7}{l}{$H_1$ significantly different from HAC-FWPD}\\
\multicolumn{7}{l}{$H_0$ statistically similar to HAC-FWPD}\\
\multicolumn{7}{l}{Best values shown in \textbf{boldface}.}\\
\end{tabular}
}
\end{center}
\end{table*}

\begin{table*}[!t]
\begin{center}
\caption{Means and standard deviations of ARI values for the HAC-FWPD algorithm against MNAR-I.}\label{mnar1HacResults}
\scriptsize{
\begin{tabular}{c c c c c c c}
\hline
Dataset & $\begin{array}{ll}
\text{HAC-}\\
\text{-FWPD}\\
\end{array}$ & ZI & MI & SVDI & kNNI & $\begin{array}{cc}
\text{Best } k\\
\text{(for kNNI)}\\
\end{array}$\\
\hline
Chronic Kidney & \textbf{1.000}$\pm$0.000 & \textbf{1.000}$\pm$0.000 & \textbf{1.000}$\pm$0.000 & \textbf{1.000}$\pm$0.000 & 0.002$\pm$0.000 & 3 \\
Colon 		   & \textbf{0.926}$\pm$0.166 & 0.025$\pm$0.000 & 0.025$\pm$0.000 & 0.025$\pm$0.000 & 0.014$\pm$0.000 & 3 \\
GSAD 1 		   & \textbf{0.473}$\pm$0.237 & 0.326$\pm$0.085 & 0.343$\pm$0.098 & 0.271$\pm$0.112 & 0.005$\pm$0.003 & 3 \\
Glass          & 0.736$\pm$0.135 & 0.697$\pm$0.119 & \textbf{0.738}$\pm$0.133 & 0.614$\pm$0.145 & 0.018$\pm$0.006 & 10 \\
Iris           & 0.852$\pm$0.142 & 0.527$\pm$0.437 & 0.543$\pm$0.456 & \textbf{0.881}$\pm$0.180 & 0.540$\pm$0.026 & 20\\
Isolet 5 	   & \textbf{0.586}$\pm$0.085 & 0.326$\pm$0.207 & 0.401$\pm$0.169 & 0.223$\pm$0.153 & 0.048$\pm$0.033 & 3 \\
Landsat 	   & \textbf{0.786}$\pm$0.085 & 0.420$\pm$0.298 & 0.443$\pm$0.375 & 0.765$\pm$0.086 & 0.072$\pm$0.004 & 5 \\
Leaf 		   & \textbf{0.514}$\pm$0.043 & 0.345$\pm$0.065 & 0.267$\pm$0.080 & 0.437$\pm$0.054 & 0.110$\pm$0.035 & 5 \\
Libras         & \textbf{0.843}$\pm$0.109 & 0.750$\pm$0.101 & 0.750$\pm$0.101 & 0.782$\pm$0.087 & 0.419$\pm$0.042 & 3 \\
Lung 		   & \textbf{1.000}$\pm$0.000 & \textbf{1.000}$\pm$0.000 & \textbf{1.000}$\pm$0.000 & \textbf{1.000}$\pm$0.000 & 0.008$\pm$0.024 & 3 \\
Lung Cancer    & \textbf{0.651}$\pm$0.316 & 0.516$\pm$0.269 & 0.516$\pm$0.269 & 0.561$\pm$0.336 & 0.020$\pm$0.010 & 3 \\
Lymphoma 	   & \textbf{0.942}$\pm$0.130 & 0.861$\pm$0.000 & 0.861$\pm$0.000 & 0.861$\pm$0.000 & 0.651$\pm$0.000 & 3 \\
Pendigits      & \textbf{0.641}$\pm$0.172 & 0.405$\pm$0.255 & 0.341$\pm$0.238 & 0.399$\pm$0.139 & 0.416$\pm$0.043 & 5 \\
Prostate 	   & \textbf{1.000}$\pm$0.000 & \textbf{1.000}$\pm$0.000 & \textbf{1.000}$\pm$0.000 & \textbf{1.000}$\pm$0.000 & 0.001$\pm$0.000 & 3 \\
Seeds          & \textbf{0.584}$\pm$0.154 & 0.429$\pm$0.141 & 0.479$\pm$0.047 & 0.421$\pm$0.084 & 0.555$\pm$0.082 & 3 \\
Sensorless 	   & \textbf{0.300}$\pm$0.298 & 0.225$\pm$0.025 & 0.217$\pm$0.027 & 0.216$\pm$0.009 & 0.000$\pm$0.000 & 3 \\
Sonar 		   & \textbf{0.598}$\pm$0.550 & 0.196$\pm$0.449 & 0.196$\pm$0.449 & 0.329$\pm$0.473 & 0.001$\pm$0.000 & 3 \\
Theorem Proving & \textbf{0.775}$\pm$0.121 & 0.762$\pm$0.068 & 0.731$\pm$0.025 & 0.742$\pm$0.031 & 0.002$\pm$0.000 & 3 \\
Vehicle        & \textbf{0.797}$\pm$0.121 & 0.321$\pm$0.194 & 0.522$\pm$0.264 & 0.699$\pm$0.290 & 0.068$\pm$0.040 & 3 \\
Vowel Context  & \textbf{0.447}$\pm$0.168 & 0.300$\pm$0.073 & 0.282$\pm$0.068 & 0.306$\pm$0.114 & 0.095$\pm$0.029 & 3 \\
\hline
Average Ranks  & \textbf{1.33} & 3.15 & 3.05 & 2.83 & 4.65 &   \\
\hline
\multicolumn{2}{c}{Signed Rank Hypotheses (p-values)} & $H_1 (0.00)$ & $H_1 (0.00)$ & $H_1 (0.00)$ & $H_1 (0.00)$ & \\
\hline
\multicolumn{7}{l}{$H_1$ significantly different from HAC-FWPD}\\
\multicolumn{7}{l}{$H_0$ statistically similar to HAC-FWPD}\\
\multicolumn{7}{l}{Best values shown in \textbf{boldface}.}\\
\end{tabular}
}
\end{center}
\end{table*}

\begin{table*}[!t]
\begin{center}
\caption{Means and standard deviations of ARI values for the HAC-FWPD algorithm against MNAR-II.}\label{mnar2HacResults}
\scriptsize{
\begin{tabular}{c c c c c c c}
\hline
Dataset & $\begin{array}{ll}
\text{HAC-}\\
\text{-FWPD}\\
\end{array}$ & ZI & MI & SVDI & kNNI & $\begin{array}{cc}
\text{Best } k\\
\text{(for kNNI)}\\
\end{array}$\\
\hline
Chronic Kidney & \textbf{1.000}$\pm$0.000 & \textbf{1.000}$\pm$0.000 & \textbf{1.000}$\pm$0.000 & \textbf{1.000}$\pm$0.000 & 0.002$\pm$0.000 & 3 \\
Colon 		   & \textbf{1.000}$\pm$0.000 & 0.147$\pm$0.384 & 0.147$\pm$0.384 & 0.147$\pm$0.384 & 0.013$\pm$0.004 & 3 \\
GSAD 1 		   & \textbf{0.429}$\pm$0.190 & 0.353$\pm$0.011 & 0.356$\pm$0.021 & 0.353$\pm$0.017 & 0.009$\pm$0.001 & 3 \\
Glass          & 0.717$\pm$0.125 & \textbf{0.733}$\pm$0.103 & 0.696$\pm$0.113 & 0.690$\pm$0.172 & 0.020$\pm$0.008 & 20\\
Iris           & 0.885$\pm$0.079 & 0.718$\pm$0.394 & 0.718$\pm$0.394 & \textbf{0.928}$\pm$0.024 & 0.552$\pm$0.012 & 10\\
Isolet 5 	   & \textbf{0.711}$\pm$0.047 & 0.296$\pm$0.214 & 0.479$\pm$0.007 & 0.184$\pm$0.174 & 0.043$\pm$0.034 & 3 \\
Landsat 	   & \textbf{0.763}$\pm$0.048 & 0.229$\pm$0.004 & 0.229$\pm$0.004 & 0.651$\pm$0.278 & 0.082$\pm$0.003 & 3 \\
Leaf 		   & \textbf{0.477}$\pm$0.036 & 0.255$\pm$0.109 & 0.277$\pm$0.081 & 0.342$\pm$0.124 & 0.110$\pm$0.044 & 3 \\
Libras         & \textbf{0.817}$\pm$0.053 & 0.742$\pm$0.057 & 0.758$\pm$0.065 & 0.780$\pm$0.097 & 0.392$\pm$0.040 & 3 \\
Lung 		   & \textbf{1.000}$\pm$0.000 & \textbf{1.000}$\pm$0.000 & \textbf{1.000}$\pm$0.000 & \textbf{1.000}$\pm$0.000 & 0.009$\pm$0.019 & 3 \\
Lung Cancer    & \textbf{0.527}$\pm$0.158 & 0.515$\pm$0.194 & 0.515$\pm$0.194 & 0.503$\pm$0.216 & 0.035$\pm$0.021 & 3 \\
Lymphoma 	   & \textbf{0.925}$\pm$0.122 & 0.916$\pm$0.076 & 0.876$\pm$0.034 & 0.876$\pm$0.034 & 0.706$\pm$0.075 & 3 \\
Pendigits      & \textbf{0.485}$\pm$0.132 & 0.219$\pm$0.053 & 0.300$\pm$0.117 & 0.427$\pm$0.108 & 0.387$\pm$0.030 & 20\\
Prostate 	   & \textbf{1.000}$\pm$0.000 & \textbf{1.000}$\pm$0.000 & \textbf{1.000}$\pm$0.000 & \textbf{1.000}$\pm$0.000 & 0.001$\pm$0.000 & 3 \\
Seeds          & \textbf{0.581}$\pm$0.154 & 0.398$\pm$0.079 & 0.319$\pm$0.149 & 0.491$\pm$0.149 & 0.580$\pm$0.125 & 3\\
Sensorless 	   & \textbf{0.475}$\pm$0.378 & 0.204$\pm$0.061 & 0.210$\pm$0.044 & 0.214$\pm$0.045 & 0.000$\pm$0.000 & 3 \\
Sonar 		   & \textbf{0.295}$\pm$0.449 & 0.196$\pm$0.450 & 0.196$\pm$0.450 & 0.261$\pm$0.436 & 0.001$\pm$0.000 & 3 \\
Theorem Proving & \textbf{0.885}$\pm$0.078 & 0.681$\pm$0.057 & 0.681$\pm$0.057 & 0.711$\pm$0.059 & 0.001$\pm$0.002 & 3 \\
Vehicle        & \textbf{0.821}$\pm$0.072 & 0.518$\pm$0.289 & 0.664$\pm$0.251 & 0.700$\pm$0.278 & 0.041$\pm$0.037 & 3 \\
Vowel Context  & \textbf{0.533}$\pm$0.154 & 0.377$\pm$0.232 & 0.430$\pm$0.109 & 0.425$\pm$0.109 & 0.103$\pm$0.024 & 3 \\
\hline
Average Ranks  & \textbf{1.33} & 3.33 & 3.00 & 2.60 & 4.75 &    \\
\hline
\multicolumn{2}{c}{Signed Rank Hypotheses (p-values)} & $H_1 (0.00)$ & $H_1 (0.00)$ & $H_1 (0.00)$ & $H_1 (0.00)$ & \\
\hline
\multicolumn{7}{l}{$H_1$ significantly different from HAC-FWPD}\\
\multicolumn{7}{l}{$H_0$ statistically similar to HAC-FWPD}\\
\multicolumn{7}{l}{Best values shown in \textbf{boldface}.}\\
\end{tabular}
}
\end{center}
\end{table*}

It is seen from Tables \ref{mcarHacResults}-\ref{mnar2HacResults} that the HAC-FWPD algorithm is able to perform best on all types of missingness, as evident from the consistently minimum average ranking.
The proposed method performs best on the majority of datasets for all types of missingness. Moreover, the performance of HAC-FWPD is observed to be significantly better than kNNI which performs poorly overall, indicating that kNNI may not be useful for hierarchical clustering applications with missingness. Interestingly, in case of MAR and MNAR-II, both of which are characterized by the dependence of the missingness on the observed features, ZI, MI as well as SVDI show improved performance. This indicates that the dependence of the missingness on the observed features aids these imputation methods in case of hierarchical clustering. Another intriguing observation is that all the contending HAC methods consistently achieved the best possible performance on the high-dimensional datasets Lung and Prostate. This may indicate that while convexity of the cluster structures may be harmed due to missingness, the local proximity among the points is preserved owing to the sparse nature of such high-dimensional datasets.

\section{Conclusions}\label{sec:conclsn}

In this paper, we propose to use the FWPD measure as a viable alternative to imputation and marginalization approaches to handle the problem of missing features in data clustering. The proposed measure attempts to estimate the original distances between the data points by adding a penalty term to those pair-wise distances which cannot be calculated on the entire feature space due to missing features. Therefore, unlike existing methods for handling missing features, FWPD is also able to distinguish between distinct data points which look identical due to missing features. Yet, FWPD also ensures that the dissimilarity for any data instance from itself is never greater than its dissimilarity from any other point in the dataset. Intuitively, these advantages of FWPD should help us better model the original data space which may help in achieving better clustering performance on the incomplete data.

Therefore, we use the proposed FWPD measure to put forth the k-means-FWPD and the HAC-FWPD clustering algorithms, which are directly applicable to datasets with missing features. We conduct extensive experimentation on the new techniques using various benchmark datasets and find the new approach to produce generally better results (for both partitional as well as hierarchical clustering) compared to some of the popular imputation methods which are generally used to handle the missing feature problem. In fact, it is observed from the experiments that the performance of the imputation schemes varies with the type of missingness and/or the clustering algorithm being used (for example, kNNI is useful for k-means clustering but not for HAC clustering; SVDI is useful for HAC clustering but not for k-means clustering; MI is effective when the missingness depends on the observed features). The proposed approach, on the other hand, exhibits good performance across all types of missingness as well as both partitional and hierarchical clustering paradigms. The experimental results attest to the ability of FWPD to better model the original data space, compared to existing methods.

However, it must be stressed, that the performance of all these methods, including the FWPD based ones, can vary depending on the structure of the dataset concerned, the choice of the proximity measure used (for HAC), and the pattern and extent of missingness plaguing the data. Fortunately, the $\alpha$ parameter embedded in FWPD can be varied in accordance with the extent of missingness to achieve desired results. The results in Section \ref{sec:alpha} indicate that it may be useful to choose a high value of $\alpha$ when a large fraction of the features are unobserved, and to choose a smaller value when only a few of the features are missing. However, in the presence of a sizable amount of missingness and the absence of ground-truths to validate the merit of the achieved clusterings, it is safest to choose a value of $\alpha$ proportional to the percentage of missing features restricted within the range $[0.1,0.25]$. We also present an appendix dealing with an extension of the FWPD measure to problems with absent features and show that this modified form of FWPD is a semi-metric. 

\par An obvious follow-up to this work is the application of the proposed PDM variant to practical clustering problems which are characterized by large fractions of unobserved data that arise in various fields such as economics, psychiatry, web-mining, etc. Studies can be undertaken to better understand the effects that the choice of $\alpha$ has on the clustering results. Another rewarding topic of research is the investigation of the abilities of the FWPD variant for absent features (see Appendix \ref{apd:first}) by conducting proper experiments using benchmark applications characterized by this rare form of missingness (structural missingness).

\appendix

\normalsize{

\section{Extending the FWPD to problems with Absent Features}\label{apd:first}

This appendix proposes an extension of the FWPD measure to the case of absent features or structural missingness. The principal difference between missing and absent features lies in the fact that the unobserved features are known to be undefined in the latter case, unlike the former. Therefore, while it makes sense to add penalties for features which are observed for only one of the data instances (as the very existence of such a feature sets the points apart), it makes little sense to add penalties for features which are undefined for both the data points. This is in contrast to problems with unstructured missingness where a feature missing from both the data instances is known to be defined for both points (which potentially have distinct values of this feature). Thus, the fundamental difference between the problems of missing and absent features is that two points observed in the same subspace and having identical observed features should (unlike the missing data problem) essentially be considered identical instances in the case of absent features, as the unobserved features are known to be non-existent. But, in case of the unobserved features being merely unknown (rather than being non-existent), such data points should be considered distinct because the unobserved features are likely to have distinct values (making the points distinct when completely observed).  Hence, it is essential to add penalties for features missing from both points in the case of missing features, but not in the case of absent features. Keeping this in mind, we can modify the proposed FWPD (essentially modifying the proposed FWP) as defined in the following text to serve as a dissimilarity measure for structural missingness.

\par Let the dataset $X_{abs}$ consist of $n$ data instances $\mathbf{x}_i$ ($i \in \{1, 2, \cdots, n\}$). Let $\zeta_{\mathbf{x}_i}$ denote the set of features on which the data point $\mathbf{x}_i \in X_{abs}$ is defined.

\begin{dfn}\label{defAbsFwp}
The FWP between the instances $\mathbf{x}_i$ and $\mathbf{x}_j$ in $X_{abs}$ is defined as
\begin{equation}\label{eqnDefAbsFwp}
p_{abs}(\mathbf{x}_i,\mathbf{x}_j)=\frac{\underset{l \in (\zeta_{\mathbf{x}_i}\bigcup \zeta_{\mathbf{x}_j}) \backslash (\zeta_{\mathbf{x}_i}\bigcap \zeta_{\mathbf{x}_j})}{\sum}\;\nu_l}{\underset{l' \in \zeta_{\mathbf{x}_i}\bigcup \zeta_{\mathbf{x}_j}}{\sum}\;\nu_{l'}}
\end{equation}
where $\nu_s \in (0,n]$ is the number of instances in $X_{abs}$ that are characterized by the feature $s$. Like in the case of unstructured missingness, this FWP also exacts greater penalty for the non-existence of commonly features.
\end{dfn}

Then, the definition of the FWPD modified for structural missingness is as follows.
\begin{dfn}\label{defAbsFwpd}
The FWPD between $\mathbf{x}_i$ and $\mathbf{x}_j$ in $X_{abs}$ is
\begin{equation}\label{eqnAbsFwpd}
\delta_{abs}(\mathbf{x}_i,\mathbf{x}_j)=(1-\alpha)\times \frac{d(\mathbf{x}_i,\mathbf{x}_j)}{d_{max}} + \alpha \times p_{abs}(\mathbf{x}_i,\mathbf{x}_j),
\end{equation}
where $\alpha \in (0,1)$ is a parameter which determines the relative importance between the two terms and $d(\mathbf{x}_i,\mathbf{x}_j)$ and $d_{max}$ retain their former definitions (but, in the context of structural missingness).
\end{dfn}

\par Now, having modified the FWPD to handle structural missingness, we show in the following theorem that the modified FWPD is a semi-metric.
\begin{theorem}\label{thmPropAbsFwpd}
The FWPD for absent features is a semi-metric, i.e. it satisfies the following important properties:
\begin{enumerate}
\item $\delta_{abs}(\mathbf{x}_i,\mathbf{x}_j) \geq 0$ $\forall$ $\mathbf{x}_i,\mathbf{x}_j \in X_{abs}$,
\item $\delta_{abs}(\mathbf{x}_i,\mathbf{x}_j) = 0$ iff $\mathbf{x}_i = \mathbf{x}_j$, i.e. $\zeta_{\mathbf{x}_i} = \zeta_{\mathbf{x}_j}$ and $x_{i,l} = x_{j,l}$ $\forall$ $l \in \zeta_{\mathbf{x}_i}$, and
\item $\delta_{abs}(\mathbf{x}_i,\mathbf{x}_j)=\delta_{abs}(\mathbf{x}_j,\mathbf{x}_i)$ $\forall$ $\mathbf{x}_i,\mathbf{x}_j \in X_{abs}$.
\end{enumerate}

\begin{proof}\label{pfPropAbsFwpd}
\begin{enumerate}
\item From Equation (\ref{eqnDefAbsFwp}) we can see that $p_{abs}(\mathbf{x}_i,\mathbf{x}_j) \geq 0$ $\forall$ $\mathbf{x}_i,\mathbf{x}_j \in X_{abs}$ and Equation (\ref{eqnLowerDist}) implies that $d(\mathbf{x}_i,\mathbf{x}_j) \geq 0$ $\forall$ $\mathbf{x}_i,\mathbf{x}_j \in X_{abs}$. Hence, it follows that $\delta_{abs}(\mathbf{x}_i,\mathbf{x}_j) \geq 0$ $\forall$ $\mathbf{x}_i,\mathbf{x}_j \in X_{abs}$.
\item It is easy to see from Equation (\ref{eqnDefAbsFwp}) that $p_{abs}(\mathbf{x}_i,\mathbf{x}_i)=0$ iff $\zeta_{\mathbf{x}_i}=\zeta_{\mathbf{x}_j}$. Now, if $x_{i,l} = x_{j,l}$ $\forall$ $l \in \zeta_{\mathbf{x}_i}$, then $d(\mathbf{x}_i,\mathbf{x}_j) = 0$. Hence, $\delta_{abs}(\mathbf{x}_i,\mathbf{x}_j) = 0$ when $\zeta_{\mathbf{x}_i} = \zeta_{\mathbf{x}_j}$ and $x_{i,l} = x_{j,l}$ $\forall$ $l \in \zeta_{\mathbf{x}_i}$. The converse is also true as $\delta_{abs}(\mathbf{x}_i,\mathbf{x}_j) = 0$ implies $\zeta_{\mathbf{x}_i} = \zeta_{\mathbf{x}_j}$ and $d(\mathbf{x}_i,\mathbf{x}_j) = 0$; the latter in turn implying that $x_{i,l} = x_{j,l}$ $\forall$ $l \in \zeta_{\mathbf{x}_i}$.
\item From Equation (\ref{eqnAbsFwpd}) we have
    \begin{equation*}
    \begin{aligned}
    & \delta_{abs}(\mathbf{x}_i,\mathbf{x}_j)=(1-\alpha)\times \frac{d(\mathbf{x}_i,\mathbf{x}_j)}{d_{max}} + \alpha \times p_{abs}(\mathbf{x}_i,\mathbf{x}_j),\\
    \text{and } & \delta_{abs}(\mathbf{x}_j,\mathbf{x}_i)=(1-\alpha)\times \frac{d(\mathbf{x}_j,\mathbf{x}_i)}{d_{max}} + \alpha \times p_{abs}(\mathbf{x}_j,\mathbf{x}_i).
    \end{aligned}
    \end{equation*}
    But, $d(\mathbf{x}_i,\mathbf{x}_j)=d(\mathbf{x}_j,\mathbf{x}_i)$ and $p_{abs}(\mathbf{x}_i,\mathbf{x}_j)=p(\mathbf{x}_j,\mathbf{x}_i)$ $\forall$ $\mathbf{x}_i,\mathbf{x}_j \in X_{abs}$ (by definition). Therefore, it can be easily seen that $\delta_{abs}(\mathbf{x}_i,\mathbf{x}_j)=\delta_{abs}(\mathbf{x}_j,\mathbf{x}_i)$ $\forall$ $\mathbf{x}_i,\mathbf{x}_j \in X_{abs}$.
\end{enumerate}
\end{proof}

\end{theorem}

}

\begin{acknowledgements}
We would like to thank Debaleena Misra and Sayak Nag, formerly of the Department of Instrumentation and Electronics Engineering, Jadavpur University, Kolkata, India, for their extensive help with the computer implementations of the different techniques used in our experiments.
\end{acknowledgements}

%


\bibliographystyle{theapa}
\bibliography{missBib1}

\begin{thebibliography}{}

\bibitem[\protect\BCAY{Acu{\~n}a\ \BBA\ Rodriguez}{Acu{\~n}a\ \BBA\
  Rodriguez}{2004}]{acuna2004treatment}
Acu{\~n}a, E.\BBACOMMA\  \BBA\ Rodriguez, C. \BBOP2004\BBCP.
\newblock \BBOQ The treatment of missing values and its effect on classifier
  accuracy\BBCQ\
\newblock In Banks, D., McMorris, F.~R., Arabie, P., \BBA\ Gaul, W.\BEDS, {\Bem
  Classification, Clustering, and Data Mining Applications}, Studies in
  Classification, Data Analysis, and Knowledge Organisation, \BPGS\ 639--647.
  Springer Berlin Heidelberg.

\bibitem[\protect\BCAY{Ahmad\ \BBA\ Tresp}{Ahmad\ \BBA\
  Tresp}{1993}]{ahmad1993some}
Ahmad, S.\BBACOMMA\  \BBA\ Tresp, V. \BBOP1993\BBCP.
\newblock \BBOQ Some solutions to the missing feature problem in vision\BBCQ\
\newblock In Hanson, S., Cowan, J., \BBA\ Giles, C.\BEDS, {\Bem Advances in
  Neural Information Processing Systems 5}, \BPGS\ 393--400. Morgan-Kaufmann.

\bibitem[\protect\BCAY{Barcel{\'o}}{Barcel{\'o}}{2008}]{barcelo2008impact}
Barcel{\'o}, C. \BBOP2008\BBCP.
\newblock \BBOQ The impact of alternative imputation methods on the measurement
  of income and wealth: Evidence from the spanish survey of household
  finances\BBCQ\
\newblock In {\Bem Working Paper Series}. Banco de Espa{\~n}a.

\bibitem[\protect\BCAY{Bo, Dysvik,\ \BBA\ Jonassen}{Bo
  et~al.}{2004}]{bo2004lsimpute}
Bo, T.~H., Dysvik, B., \BBA\ Jonassen, I. \BBOP2004\BBCP.
\newblock \BBOQ Lsimpute: accurate estimation of missing values in microarray
  data with least squares methods\BBCQ\
\newblock {\Bem Nucleic Acid Research}, {\Bem 32\/}(3).

\bibitem[\protect\BCAY{Broder, Glassman, Manasse,\ \BBA\ Zweig}{Broder
  et~al.}{1997}]{broder1997syntactic}
Broder, A.~Z., Glassman, S.~C., Manasse, M.~S., \BBA\ Zweig, G. \BBOP1997\BBCP.
\newblock \BBOQ Syntactic clustering of the web\BBCQ\
\newblock {\Bem Computer Networks and ISDN Systems}, {\Bem 29\/}(8-13),
  1157--1166.

\bibitem[\protect\BCAY{Chan\ \BBA\ Dunn}{Chan\ \BBA\
  Dunn}{1972}]{chan1972oldest}
Chan, L.~S.\BBACOMMA\  \BBA\ Dunn, O.~J. \BBOP1972\BBCP.
\newblock \BBOQ The treatment of missing values in discriminant analysis-1. the
  sampling experiment\BBCQ\
\newblock {\Bem Journal of the American Statistical Association}, {\Bem
  67\/}(338), 473--477.

\bibitem[\protect\BCAY{Chaturvedi, Carroll, Green,\ \BBA\ Rotondo}{Chaturvedi
  et~al.}{1997}]{chaturvedi1997feature}
Chaturvedi, A., Carroll, J.~D., Green, P.~E., \BBA\ Rotondo, J.~A.
  \BBOP1997\BBCP.
\newblock \BBOQ A feature-based approach to market segmentation via overlapping
  k-centroids clustering\BBCQ\
\newblock {\Bem Journal of Marketing Research}, 370--377.

\bibitem[\protect\BCAY{Chechik, Heitz, Elidan, Abbeel,\ \BBA\ Koller}{Chechik
  et~al.}{2008}]{chechik2008absent}
Chechik, G., Heitz, G., Elidan, G., Abbeel, P., \BBA\ Koller, D.
  \BBOP2008\BBCP.
\newblock \BBOQ Max-margin classification of data with absent features\BBCQ\
\newblock {\Bem Journal of Machine Learning Research}, {\Bem 9}, 1--21.

\bibitem[\protect\BCAY{Chen}{Chen}{2013}]{chen2013nomore}
Chen, F. \BBOP2013\BBCP.
\newblock \BBOQ Missing no more: Using the mcmc procedure to model missing
  data\BBCQ\
\newblock In {\Bem Proceedings of the SAS Global Forum 2013 Conference}, \BPGS\
  1--23. SAS Institute Inc.

\bibitem[\protect\BCAY{Datta, Bhattacharjee,\ \BBA\ Das}{Datta
  et~al.}{2016a}]{DattaBD16}
Datta, S., Bhattacharjee, S., \BBA\ Das, S. \BBOP2016a\BBCP.
\newblock \BBOQ Clustering with missing features: {A} penalized dissimilarity
  measure based approach\BBCQ\
\newblock {\Bem CoRR}, {\Bem abs/1604.06602}.

\bibitem[\protect\BCAY{Datta, Misra,\ \BBA\ Das}{Datta
  et~al.}{2016b}]{datta2016fwpd}
Datta, S., Misra, D., \BBA\ Das, S. \BBOP2016b\BBCP.
\newblock \BBOQ A feature weighted penalty based dissimilarity measure for
  k-nearest neighbor classification with missing features\BBCQ\
\newblock {\Bem Pattern Recognition Letters}, {\Bem 80}, 231--237.

\bibitem[\protect\BCAY{Dempster\ \BBA\ Rubin}{Dempster\ \BBA\
  Rubin}{1983}]{dempster1983incomplete}
Dempster, A.~P.\BBACOMMA\  \BBA\ Rubin, D.~B. \BBOP1983\BBCP.
\newblock {\Bem Part I: Introduction}, \lowercase{\BVOL}~2, \BPGS\ 3--10.
\newblock New York: Academic Press.

\bibitem[\protect\BCAY{Dheeru\ \BBA\ Karra~Taniskidou}{Dheeru\ \BBA\
  Karra~Taniskidou}{2017}]{dheeru2017uci}
Dheeru, D.\BBACOMMA\  \BBA\ Karra~Taniskidou, E. \BBOP2017\BBCP.
\newblock \BBOQ {UCI} machine learning repository\BBCQ\
\newblock Online Repository.

\bibitem[\protect\BCAY{Dixon}{Dixon}{1979}]{dixon1979pattern}
Dixon, J.~K. \BBOP1979\BBCP.
\newblock \BBOQ Pattern recognition with partly missing data\BBCQ\
\newblock {\Bem Systems, Man and Cybernetics, IEEE Transactions on}, {\Bem
  9\/}(10), 617--621.

\bibitem[\protect\BCAY{Donders, van~der Heijden, Stijnen,\ \BBA\ Moons}{Donders
  et~al.}{2006}]{donders2006review}
Donders, A. R.~T., van~der Heijden, G. J. M.~G., Stijnen, T., \BBA\ Moons, K.
  G.~M. \BBOP2006\BBCP.
\newblock \BBOQ Review: A gentle introduction to imputation of missing
  values\BBCQ\
\newblock {\Bem Journal of Clinical Epidemiology}, {\Bem 59\/}(10), 1087--1091.

\bibitem[\protect\BCAY{Forgy}{Forgy}{1965}]{forgy1965cluster}
Forgy, E.~W. \BBOP1965\BBCP.
\newblock \BBOQ Cluster analysis of multivariate data: Efficiency versus
  interpretability of classifications\BBCQ\
\newblock {\Bem Biometrics}, {\Bem 21}, 768--769.

\bibitem[\protect\BCAY{Grzymala-Busse\ \BBA\ Hu}{Grzymala-Busse\ \BBA\
  Hu}{2001}]{grzymala2001comparison}
Grzymala-Busse, J.~W.\BBACOMMA\  \BBA\ Hu, M. \BBOP2001\BBCP.
\newblock \BBOQ A comparison of several approaches to missing attribute values
  in data mining\BBCQ\
\newblock In {\Bem Rough Sets and Current Trends in Computing}, \BPGS\
  378--385. Springer.

\bibitem[\protect\BCAY{Hathaway\ \BBA\ Bezdek}{Hathaway\ \BBA\
  Bezdek}{2001}]{hathaway2001fcm}
Hathaway, R.~J.\BBACOMMA\  \BBA\ Bezdek, J.~C. \BBOP2001\BBCP.
\newblock \BBOQ Fuzzy c-means clustering of incomplete data\BBCQ\
\newblock {\Bem Systems, Man, and Cybernetics: Part B: Cybernetics, IEEE
  Transactions on}, {\Bem 31\/}(5), 735--744.

\bibitem[\protect\BCAY{Haveliwala, Gionis,\ \BBA\ Indyk}{Haveliwala
  et~al.}{2000}]{haveliwala2000scalable}
Haveliwala, T., Gionis, A., \BBA\ Indyk, P. \BBOP2000\BBCP.
\newblock
\newblock \BBOQ Scalable techniques for clustering the web\BBCQ.

\bibitem[\protect\BCAY{Heitjan\ \BBA\ Basu}{Heitjan\ \BBA\
  Basu}{1996}]{heitjan1996distinguishing}
Heitjan, D.~F.\BBACOMMA\  \BBA\ Basu, S. \BBOP1996\BBCP.
\newblock \BBOQ Distinguishing "missing at random" and "missing completely at
  random"\BBCQ\
\newblock {\Bem The American Statistician}, {\Bem 50\/}(3), 207--213.

\bibitem[\protect\BCAY{Himmelspach\ \BBA\ Conrad}{Himmelspach\ \BBA\
  Conrad}{2010}]{himmelspach2010clustering}
Himmelspach, L.\BBACOMMA\  \BBA\ Conrad, S. \BBOP2010\BBCP.
\newblock \BBOQ Clustering approaches for data with missing values: Comparison
  and evaluation\BBCQ\
\newblock In {\Bem Digital Information Management (ICDIM), 2010 Fifth
  International Conference on}, \BPGS\ 19--28.

\bibitem[\protect\BCAY{Horton\ \BBA\ Lipsitz}{Horton\ \BBA\
  Lipsitz}{2001}]{horton2001multiple}
Horton, N.~J.\BBACOMMA\  \BBA\ Lipsitz, S.~R. \BBOP2001\BBCP.
\newblock \BBOQ Multiple imputation in practice: Comparison of software
  packages for regression models with missing variables\BBCQ\
\newblock {\Bem The American Statistician}, {\Bem 55\/}(3), 244--254.

\bibitem[\protect\BCAY{Hubert\ \BBA\ Arabie}{Hubert\ \BBA\
  Arabie}{1985}]{hubert1985comparing}
Hubert, L.\BBACOMMA\  \BBA\ Arabie, P. \BBOP1985\BBCP.
\newblock \BBOQ Comparing partitions\BBCQ\
\newblock {\Bem Journal of Classification}, {\Bem 2\/}(1), 193--218.

\bibitem[\protect\BCAY{Jin}{Jin}{2017}]{jin2017data}
Jin, J. \BBOP2017\BBCP.
\newblock \BBOQ Genomics dataset repository\BBCQ\
\newblock Online Repository.

\bibitem[\protect\BCAY{Juszczak\ \BBA\ Duin}{Juszczak\ \BBA\
  Duin}{2004}]{juszczak2004combining}
Juszczak, P.\BBACOMMA\  \BBA\ Duin, R. P.~W. \BBOP2004\BBCP.
\newblock \BBOQ Combining one-class classifiers to classify missing data\BBCQ\
\newblock In {\Bem Multiple Classifier Systems}, \BPGS\ 92--101. Springer.

\bibitem[\protect\BCAY{Krause\ \BBA\ Polikar}{Krause\ \BBA\
  Polikar}{2003}]{krause2003ensemble}
Krause, S.\BBACOMMA\  \BBA\ Polikar, R. \BBOP2003\BBCP.
\newblock \BBOQ An ensemble of classifiers approach for the missing feature
  problem\BBCQ\
\newblock In {\Bem Proceedings of the International Joint Conference on Neural
  Networks, 2003}, \lowercase{\BVOL}~1, \BPGS\ 553--558. IEEE.

\bibitem[\protect\BCAY{Lasdon}{Lasdon}{2013}]{lasdon2013optimization}
Lasdon, L.~S. \BBOP2013\BBCP.
\newblock {\Bem Optimization theory for Large Systems}.
\newblock Courier Corporation.

\bibitem[\protect\BCAY{Lei}{Lei}{2010}]{lei2010identify}
Lei, L. \BBOP2010\BBCP.
\newblock \BBOQ Identify earthquake hot spots with 3-dimensional density-based
  clustering analysis\BBCQ\
\newblock In {\Bem Geoscience and Remote Sensing Symposium (IGARSS), 2010 IEEE
  International}, \BPGS\ 530--533. IEEE.

\bibitem[\protect\BCAY{Little\ \BBA\ Rubin}{Little\ \BBA\
  Rubin}{1987}]{little1987statistical}
Little, R. J.~A.\BBACOMMA\  \BBA\ Rubin, D.~B. \BBOP1987\BBCP.
\newblock {\Bem Statistical Analysis with Missing Data}.
\newblock John Wiley \& Sons, Inc., New York.

\bibitem[\protect\BCAY{Lloyd}{Lloyd}{1982}]{lloyd1982least}
Lloyd, S.~P. \BBOP1982\BBCP.
\newblock \BBOQ Least squares quantization in pcm\BBCQ\
\newblock {\Bem Information Theory, IEEE Transactions on}, {\Bem 28\/}(2),
  129--137.

\bibitem[\protect\BCAY{MacQueen}{MacQueen}{1967}]{macqueen1967some}
MacQueen, J. \BBOP1967\BBCP.
\newblock \BBOQ Some methods for classification and analysis of multivariate
  observations\BBCQ\
\newblock In {\Bem Proceedings of the Fifth Berkeley Symposium on Mathematical
  Statistics and Probability}, \lowercase{\BVOL}~1, \BPGS\ 281--297. University
  of California Press.

\bibitem[\protect\BCAY{Marlin}{Marlin}{2008}]{marlin2008missing}
Marlin, B.~M. \BBOP2008\BBCP.
\newblock {\Bem Missing Data Problems in Machine Learning}.
\newblock Ph.D.\ thesis, University of Toronto.

\bibitem[\protect\BCAY{Mill{\'a}n-Giraldo, Duin,\ \BBA\
  S{\'a}nchez}{Mill{\'a}n-Giraldo et~al.}{2010}]{millan2010dissimilarity}
Mill{\'a}n-Giraldo, M., Duin, R.~P., \BBA\ S{\'a}nchez, J.~S. \BBOP2010\BBCP.
\newblock \BBOQ Dissimilarity-based classification of data with missing
  attributes\BBCQ\
\newblock In {\Bem Cognitive Information Processing (CIP), 2010 2nd
  International Workshop on}, \BPGS\ 293--298. IEEE.

\bibitem[\protect\BCAY{Murtagh\ \BBA\ Contreras}{Murtagh\ \BBA\
  Contreras}{2012}]{murtagh2012algorithms}
Murtagh, F.\BBACOMMA\  \BBA\ Contreras, P. \BBOP2012\BBCP.
\newblock \BBOQ Algorithms for hierarchical clustering: an overview\BBCQ\
\newblock {\Bem Wiley Interdisciplinary Reviews: Data Mining and Knowledge
  Discovery}, {\Bem 2\/}(1), 86--97.

\bibitem[\protect\BCAY{Myrtveit, Stensrud,\ \BBA\ Olsson}{Myrtveit
  et~al.}{2001}]{myrtveit2001analyzing}
Myrtveit, I., Stensrud, E., \BBA\ Olsson, U.~H. \BBOP2001\BBCP.
\newblock \BBOQ Analyzing data sets with missing data: An empirical evaluation
  of imputation methods and likelihood-based methods\BBCQ\
\newblock {\Bem Software Engineering, IEEE Transactions on}, {\Bem 27\/}(11),
  999--1013.

\bibitem[\protect\BCAY{Nanni, Lumini,\ \BBA\ Brahnam}{Nanni
  et~al.}{2012}]{nanni2012classifier}
Nanni, L., Lumini, A., \BBA\ Brahnam, S. \BBOP2012\BBCP.
\newblock \BBOQ A classifier ensemble approach for the missing feature
  problem\BBCQ\
\newblock {\Bem Artificial Intelligence in Medicine}, {\Bem 55\/}(1), 37--50.

\bibitem[\protect\BCAY{Porro-Mu{\~n}oz, Duin,\ \BBA\ Talavera}{Porro-Mu{\~n}oz
  et~al.}{2013}]{porro2013missing}
Porro-Mu{\~n}oz, D., Duin, R.~P., \BBA\ Talavera, I. \BBOP2013\BBCP.
\newblock \BBOQ Missing values in dissimilarity-based classification of
  multi-way data\BBCQ\
\newblock In {\Bem Iberoamerican Congress on Pattern Recognition}, \BPGS\
  214--221. Springer.

\bibitem[\protect\BCAY{Rubin}{Rubin}{1976}]{rubin1976inference}
Rubin, D.~B. \BBOP1976\BBCP.
\newblock \BBOQ Inference and missing data\BBCQ\
\newblock {\Bem Biometrika}, {\Bem 63\/}(3), 581--592.

\bibitem[\protect\BCAY{Rubin}{Rubin}{1987}]{rubin1987multiple}
Rubin, D.~B. \BBOP1987\BBCP.
\newblock {\Bem Multiple Imputation for Nonresponse in Surveys}.
\newblock John Wiley \& Sons.

\bibitem[\protect\BCAY{Sabau}{Sabau}{2012}]{sabau2012survey}
Sabau, A.~S. \BBOP2012\BBCP.
\newblock \BBOQ Survey of clustering based financial fraud detection
  research\BBCQ\
\newblock {\Bem Informatica Economica}, {\Bem 16\/}(1), 110.

\bibitem[\protect\BCAY{Schafer}{Schafer}{1997}]{schafer1997analysis}
Schafer, J.~L. \BBOP1997\BBCP.
\newblock {\Bem Analysis of Incomplete Multivariate Data}.
\newblock CRC Press.

\bibitem[\protect\BCAY{Schafer\ \BBA\ Graham}{Schafer\ \BBA\
  Graham}{2002}]{schafer2002missing}
Schafer, J.~L.\BBACOMMA\  \BBA\ Graham, J.~W. \BBOP2002\BBCP.
\newblock \BBOQ Missing data: Our view of the state of the art\BBCQ\
\newblock {\Bem Psychological Methods}, {\Bem 7\/}(2), 147--177.

\bibitem[\protect\BCAY{Sehgal, Gondal,\ \BBA\ Dooley}{Sehgal
  et~al.}{2005}]{sehgal2005collateral}
Sehgal, M. S.~B., Gondal, I., \BBA\ Dooley, L.~S. \BBOP2005\BBCP.
\newblock \BBOQ Collateral missing value imputation: a new robust missing value
  estimation algorithm fpr microarray data\BBCQ\
\newblock {\Bem Bioinformatics}, {\Bem 21\/}(10), 2417--2423.

\bibitem[\protect\BCAY{Selim\ \BBA\ Ismail}{Selim\ \BBA\
  Ismail}{1984}]{selim1984kmeans}
Selim, S.~Z.\BBACOMMA\  \BBA\ Ismail, M.~A. \BBOP1984\BBCP.
\newblock \BBOQ K-means-type algorithms: A generalized convergence theorem and
  characterization of local optimality\BBCQ\
\newblock {\Bem Pattern Analysis and Machine Intelligence, IEEE Transactions
  on}, {\Bem 6\/}(1), 81--87.

\bibitem[\protect\BCAY{Shelly, Ellsworth, Ryberg, Haberland, Fuis, Murphy,
  Nadeau,\ \BBA\ B{\"u}rgmann}{Shelly et~al.}{2009}]{shelly2009precise}
Shelly, D.~R., Ellsworth, W.~L., Ryberg, T., Haberland, C., Fuis, G.~S.,
  Murphy, J., Nadeau, R.~M., \BBA\ B{\"u}rgmann, R. \BBOP2009\BBCP.
\newblock \BBOQ Precise location of san andreas fault tremors near cholame,
  california using seismometer clusters: Slip on the deep extension of the
  fault?\BBCQ\
\newblock {\Bem Geophysical research letters}, {\Bem 36\/}(1).

\bibitem[\protect\BCAY{Troyanskaya, Cantor, Sherlock, Brown, Hastie,
  Tibshirani, Botstein,\ \BBA\ Altman}{Troyanskaya
  et~al.}{2001}]{troyanskaya2001missing}
Troyanskaya, O., Cantor, M., Sherlock, G., Brown, P., Hastie, T., Tibshirani,
  R., Botstein, D., \BBA\ Altman, R.~B. \BBOP2001\BBCP.
\newblock \BBOQ Missing value estimation methods for dna microarrays\BBCQ\
\newblock {\Bem Bioinformatics}, {\Bem 17\/}(6), 520--525.

\bibitem[\protect\BCAY{Wagstaff}{Wagstaff}{2004}]{wagstaff2004clustering}
Wagstaff, K.~L. \BBOP2004\BBCP.
\newblock \BBOQ Clustering with missing values: No imputation required\BBCQ\
\newblock In {\Bem Proceedings of the Meeting of the International Federation
  of Classification Societies}, \BPGS\ 649–--658.

\bibitem[\protect\BCAY{Wagstaff\ \BBA\ Laidler}{Wagstaff\ \BBA\
  Laidler}{2005}]{wagstaff2005making}
Wagstaff, K.~L.\BBACOMMA\  \BBA\ Laidler, V.~G. \BBOP2005\BBCP.
\newblock \BBOQ Making the most of missing values: Object clustering with
  partial data in astronomy\BBCQ\
\newblock In {\Bem Astronomical Data Analysis Software and Systems XIV}, ASP
  Conference Series, \BPGS\ 172--176. Astronomical Society of the Pacific.

\bibitem[\protect\BCAY{Wang\ \BBA\ Rao}{Wang\ \BBA\
  Rao}{2002a}]{wang2002empirical1}
Wang, Q.\BBACOMMA\  \BBA\ Rao, J. N.~K. \BBOP2002a\BBCP.
\newblock \BBOQ Empirical likelihood-based inference in linear models with
  missing data\BBCQ\
\newblock {\Bem Scandinavian Journal of Statistics}, {\Bem 29\/}(3), 563--576.

\bibitem[\protect\BCAY{Wang\ \BBA\ Rao}{Wang\ \BBA\
  Rao}{2002b}]{wang2002empirical2}
Wang, Q.\BBACOMMA\  \BBA\ Rao, J. N.~K. \BBOP2002b\BBCP.
\newblock \BBOQ Empirical likelihood-based inference under imputation for
  missing response data\BBCQ\
\newblock {\Bem The Annals of Statistics}, {\Bem 30\/}(3), 896--924.

\bibitem[\protect\BCAY{Weatherill\ \BBA\ Burton}{Weatherill\ \BBA\
  Burton}{2009}]{weatherill2009delineation}
Weatherill, G.\BBACOMMA\  \BBA\ Burton, P.~W. \BBOP2009\BBCP.
\newblock \BBOQ Delineation of shallow seismic source zones using k-means
  cluster analysis, with application to the aegean region\BBCQ\
\newblock {\Bem Geophysical Journal International}, {\Bem 176\/}(2), 565--588.

\bibitem[\protect\BCAY{Wendel\ \BBA\ Hurter~Jr.}{Wendel\ \BBA\
  Hurter~Jr.}{1976}]{wendel1976minimization}
Wendel, R.~E.\BBACOMMA\  \BBA\ Hurter~Jr., A.~P. \BBOP1976\BBCP.
\newblock \BBOQ Minimization of a non-separable objective function subject to
  disjoint constraints\BBCQ\
\newblock {\Bem Operations Research}, {\Bem 24}, 643--657.

\bibitem[\protect\BCAY{Wilcoxon}{Wilcoxon}{1945}]{wilcoxon1945individual}
Wilcoxon, F. \BBOP1945\BBCP.
\newblock \BBOQ Individual comparisons by ranking methods\BBCQ\
\newblock {\Bem Biometrics bulletin}, {\Bem 1\/}(6), 80--83.

\bibitem[\protect\BCAY{Zhang, Yang,\ \BBA\ Wang}{Zhang
  et~al.}{2012}]{zhang2012software}
Zhang, W., Yang, Y., \BBA\ Wang, Q. \BBOP2012\BBCP.
\newblock \BBOQ A comparative study of absent features and unobserved values in
  software effort data\BBCQ\
\newblock {\Bem International Journal of Software Engineering and Knowledge
  Engineering}, {\Bem 22\/}(02), 185--202.

\end{thebibliography}


\end{document}